\documentclass{article} %
\usepackage{iclr2027_conference,times}

\usepackage{amsmath,amsfonts,bm}

\def\eqref#1{equation~\ref{#1}}

\def\1{\bm{1}}

\DeclareMathAlphabet{\mathsfit}{\encodingdefault}{\sfdefault}{m}{sl}
\SetMathAlphabet{\mathsfit}{bold}{\encodingdefault}{\sfdefault}{bx}{n}

\usepackage{xcolor}
\usepackage{url}
\usepackage{graphicx}
\usepackage{amsmath, amssymb}
\usepackage{multirow}
\usepackage{booktabs}
\usepackage{caption}
\usepackage{wrapfig}
\definecolor{paperlink}{HTML}{0057B8}
\usepackage[
  breaklinks=true,
  colorlinks=true,
  linkcolor=paperlink,
  citecolor=paperlink,
  urlcolor=paperlink
]{hyperref}
\usepackage[normalem]{ulem}
\definecolor{yccolor}{HTML}{FF6A00}
\definecolor{xuncolor}{HTML}{6A3DFF}
\definecolor{jhcolor}{HTML}{00A651}

\title{Can Video World Models Track \\Unobserved World States?}
\author{
 \hspace{-.8mm}\bf Joonghyuk Shin$^{1}$\quad
\bf Yicong Hong$^{2}$\quad
\bf Jaesik Park$^{1}$\quad
\bf Xun Huang$^{2}$ \\ 
\normalfont $^1$Seoul National University, $^2$Roblox
}

\iclrfinalcopy %
\begin{document}
\maketitle

\begin{abstract}
Video world models are increasingly used as simulators, but visual fidelity alone does not show that a model maintains the hidden state of the world.
We examine this difference with an action-conditioned video Shell Game, a visual analogue of $S_5$ state tracking that separates visual rendering from compositing the unobserved world state.
Trained on 5-swap chains, standard backbones (\textit{e.g.,} bidirectional and autoregressive Transformers, Mamba, and linear attention) render plausible videos and predict the correct ball location up to 5 swaps. 
However, they fail to learn the rule and generalize to longer swap chains, even with more denoising steps.
As the pixel-based diffusion loss does not force the generated frames to hold the unseen ball position, output tokens cannot carry it, and the state has to live within the architecture.
In a causal Transformer, this implicit state is an append-only KV cache, which is written once and never revised, so the model must re-compose the swaps at every chunk.
Tracking $S_5$ this way requires depth to grow with sequence length, which no fixed-depth Transformer provides.
We study what enables learning the rule, and find that length generalization requires a revisable state carried across chunks and an update expressive enough to apply a swap.
Linear attention can achieve this by allowing negative transition eigenvalues, and autoregressive Transformers can do so with nonlinear TTT fast weights (\textit{e.g.,} SwiGLU) whose online updates change the feature map used to read their state.
We further examine Memory Maze and Block World, where the state is not fixed by the input action stream alone and must be corrected from observations or keeps changing out of view, and discuss the implications for building stateful video world models.
\end{abstract}

\section{Introduction}

Recent video world models are often judged by whether their generated videos look plausible~\citep{Bruce2024GenieGI, parkerholder2024genie2}.
For simulation and agentic uses, plausibility alone is not enough.
We also need to track what the current frame does not show, such as the velocity or acceleration of objects, the positions of occluded objects, and changes that are made outside of the camera's view.
These dependences can matter for future observations even when the current frame looks the same.
As video models progress to world models, learning to track changes in unobserved states is becoming an important research question~\citep{Xiao2025WORLDMEMLC, Lillemark2026FlowEW}.

Recent work in state-tracking theory gives us a useful starting point~\citep{Merrill2024TheIO}.
Under log-precision assumptions~\citep{Merrill2022ThePT}, fixed-depth Transformers, standard state-space models, and several linear-attention variants can be simulated in $\mathrm{TC}^0$\footnote{$\mathrm{TC}^0$ is the class of problems solved by constant-depth, polynomial-size threshold circuits (a fixed parallel budget independent of input length), contained in $\mathrm{NC}^1$ and conjectured to be strictly smaller.}.
Composing permutations in the non-solvable group $S_5$ is $\mathrm{NC}^1$-hard\footnote{$\mathrm{NC}^1$ is the class of problems solved by polynomial-size, $O(\log n)$-depth Boolean circuits of bounded fan-in; the word problem of such a group ($S_5$) is $\mathrm{NC}^1$-complete~\citep{Merrill2024TheIO}.}, so these models cannot solve it at arbitrary lengths if $\mathrm{TC}^0\ne\mathrm{NC}^1$.
A recurrent model can apply the next transition to the state it has already computed, provided that its update can express the required operation~\citep{Grazzi2024UnlockingSI}.
Transformers can also work through these transitions by writing intermediate results into chain-of-thought (CoT) tokens~\citep{Merrill2024TheEP}.
These results clarify how models can represent sequential state updates, but whether video models learn such updates from visual supervision remains unclear.

We study this question with an action-conditioned video Shell Game, a visual analogue of $S_5$ state tracking.
The model sees the ball only at the start and at the final reveal, and intermediate frames show the cups being swapped without indicating which one holds it.
Standard autoregressive (AR) backbones learn to render cup motions and predict the correct ball positions at the training length, but fail on longer swap chains while still rendering plausible videos.
In language models, free-form CoT tokens can hold intermediate states and add effective depth with each generated token~\citep{Merrill2024TheEP}.
Video models' output  tokens could in principle also carry such states, but a reconstruction-only objective pins them to what is visible and our probes recover little hidden-ball information from the attention readouts. 
The KV cache keeps only a visual history of the earlier swaps and never revises its entries, so each new chunk must compose them again at a fixed depth.

The models that extrapolate in our experiments \emph{revise a carried state in place} with an update \emph{expressive enough to apply a swap}.
We find this in two settings: DeltaNet-style linear attention and nonlinear fast weights.
For DeltaNet, the recurrent state is already present and allowing negative transition eigenvalues gives the update the reflection-like operations needed for a swap~\citep{Grazzi2024UnlockingSI}.
For attention backbones, we use LaCT~\citep{Zhang2025TestTimeTD} to carry state in the weights of a small SwiGLU network ($W_1[\mathrm{silu}(W_0x)\odot(W_2x)]$), whose feature map $(W_0,W_2)$ and readout matrix $(W_1)$ are updated between chunks by test-time training (TTT).
Building on recent work~\citep{Liu2026TestTimeTW}, we find through controlled ablations that TTT-KV binding (TTT-KVB) can provide more expressive state updates when its online updates reach the nonlinear feature map.
We also, for the first time, train a video model with TTT-KVB from scratch and show promising performance on memory retrieval and state tracking. 
Our contributions are as follows:

\begin{itemize}
    \item We show that current video model backbones supervised solely on visible content cannot carry unobserved states, which makes recent discussions on state tracking directly relevant to how the state is implicitly carried across chunks and how expressive its transition is.
    \item On the Shell Game, we find two cases that extrapolate: (1) a recurrent transition widened to admit negative eigenvalues and (2) a TTT fast weight whose online nonlinear feature-map update makes its effective kernel history-dependent, and analyze their mechanisms.
    \item We discuss the implications for more general settings such as static and dynamic world exploration, where state must also be corrected from observations, and what remains open.
\end{itemize}

\section{Related Work}
\noindent\textbf{Video World Models.}
Recent video world models learn to predict future frames from past observations and actions, and now serve tasks as varied as interactive applications~\citep{parkerholder2024genie2,Zhang2025MatrixGameIW, He2025MatrixGame2A, Shin2025MotionStreamRV, Hong2025RELICIV, Tang2025HunyuanGameCraft2II, Wang2026MatrixGame3R,Seo2026GroundingWS}, robot planning~\citep{Agarwal2025CosmosWF, Kim2026CosmosPF, Gao2026DreamDojoAG, Ye2026WorldAM}, and autonomous driving~\citep{Ren2025CosmosDriveDreamsSS, Basant2026NVIDIAOR}.

Architecturally, these models range from bidirectional Transformers~\citep{Peebles2022ScalableDM} to causal Transformers with KV-cache inference~\citep{Yin2024FromSB}, Mamba~\citep{Po2025LongContextSV,Savov2025StateSpaceDiffuserBL}, and more recent DeltaNet-based backbones~\citep{Zhu2026SANAWMEM,Zhao2026SANAStreamingRS}.
They are typically trained with diffusion or flow-matching losses~\citep{Ho2020DenoisingDP,Lipman2022FlowMF,Liu2022FlowSA}, often combined with few-step distillation~\citep{Yin2023OneStepDW,Song2023ConsistencyM}, and conditioned on diverse action signals such as keyboard, camera pose, or mouse drag, frequently alongside text prompts.
Self Forcing~\citep{Huang2025SelfFB} distills a few-step autoregressive model using its own generated history, but detaches the cached earlier chunks so gradients do not pass back through the rollout (Appendix~\ref{app:no-scratchpad}).
Building on this common recipe, much recent work focuses on keeping the generated world consistent, both as stable 3D structure~\citep{Ren2025Gen3C3W,Wu2025GeometryFM} and as long-context memory~\citep{Wu2025VideoWM}.

Yet as these models scale to complex real-world scenarios and simulators, it becomes increasingly hard to handle moving objects under partial observation~\citep{Lillemark2026FlowEW,Duan2026LiveWorldSO,Ma2026OutOS}, simulate precise physics~\citep{Kang2024HowFI,Aira2024MotionCraftPZ,Gillman2025ForcePV}, or maintain a coherent world state across multiple agents~\citep{Savva2026SolarisBA,Liu2026GammaWorldGM}.

\noindent\textbf{Linear Attention, Fast Weights, and Test-Time Training.}
Efficient sequence models replace the growing key-value cache of softmax attention with a fixed-size recurrent state written and read online.
Causal linear attention ($S_t = S_{t-1} + k_t v_t^\top$, read as $o_t = S_t^\top q_t$) stores key-value associations in an additive matrix state~\citep{Schlag2021LinearTA}.
The delta rule corrects an existing association using its prediction residual (Eq.~\ref{eq:deltanet-update}), with DeltaNet providing efficient chunk-parallel training~\citep{Yang2024ParallelizingLT} and gated variants adding data-dependent decay~\citep{Yang2024GatedDN, Hatamizadeh2026GatedDD}.
Similar recurrence underlies many recent linear-attention backbones~\citep{Dao2024TransformersAS, Zhang2025KimiLA}, differing mainly in their state transition and read kernel.

Test-time training (TTT) makes the recurrent state the weights of a small inner network updated online during the forward pass, either by a key-value binding loss that associates projected keys with values (TTT-KVB,~\citet{Sun2024LearningT, Behrouz2024TitansLT,Dalal2025OneMinuteVG}) or by backpropagating the outer task loss end-to-end through the recurrence onto a subset of the weights (TTT-E2E,~\citet{Tandon2025EndtoEndTT}). 
LaCT~\citep{Zhang2025TestTimeTD} pairs local attention with nonlinear fast weights updated over large chunks for hardware efficiency.
When the inner model is a single linear layer, TTT-KVB recovers linear attention under the dot-product loss and DeltaNet under the MSE loss~\citep{Sun2024LearningT,Liu2026TestTimeTW} (Appendix~\ref{app:ttt-linear-attention}).

\noindent\textbf{State Tracking and Expressivity.}
The $S_5$ word problem is a standard test of whether a model can compose sequential state updates, and fixed-depth Transformers and state-space models cannot solve it at arbitrary lengths~\citep{Merrill2022ThePT, Merrill2024TheIO}.
Prior work addresses this limitation in two ways.
The first changes the recurrent update.
For a linear RNN $h_t=A(x_t)h_{t-1}+b(x_t)$, the eigenvalues of the transition $A(x_t)$ decide which operations the model can track.
\citet{Grazzi2024UnlockingSI} extend this range to $[-1,1]$ so that the transition can represent reflections, which is enough for parity, and \citet{Siems2025DeltaProductIS} compose several reflections per token to obtain rotations, modular counting, and longer permutations.
Nonlinear recurrences can express more general updates but lose parallel-scan training, and several recent works study this trade-off~\citep{Siems2026LearningSF, Schne2025ImplicitLM, Merrill2026WhyAL,Mishra2026M2RNNNR}.
The second adds serial computation at inference, either by generating chain-of-thought tokens or by running the backbone repeatedly in looped and recursive models~\citep{Dehghani2018UniversalT, Giannou2023LoopedTA, Saunshi2025ReasoningWL, Geiping2025ScalingUT, Wang2025HierarchicalRM, jolicoeurmartineau2025morerecursivereasoningtiny}.

This line of analysis has mostly focused on language, but recent work has begun to study the visual domain.
For generation, \citet{Ke2025CircuitCB} place visual autoregressive models within $\mathrm{TC}^0$, and \citet{Liu2025TheSS} and \citet{Chao2026TheSG} show that iterative denoising adds no serial depth of its own.
For understanding, recent benchmarks find video models far below humans at visual state tracking~\citep{Liu2026CanVM, Yu2026BenchmarkingVS}.
In this study, we ask whether an action-conditioned video model can learn to track an unobserved state from a pixel-level supervised objective alone, which is also an important question about whether video diffusion training is a scalable path toward world models.

\section{From \texorpdfstring{$S_5$}{S5} to the Visual Shell Game}
\label{sec:benchmark}

\subsection{Compositional State Tracking}
\label{sec:formal}
We use the finite-group word problem to study compositional state tracking.
Given group elements $g_1,g_2,\ldots,g_t$, the model must predict their product $g_1g_2\cdots g_t$.
Our tasks use $S_3$ and $S_5$, the permutation groups on three and five elements.
$S_3$ is solvable, while $S_5$ is the smallest non-solvable symmetric group.
Its word problem is $\mathrm{NC}^1$-complete~\citep{Barrington1986BoundedwidthPB}, making it a useful test for constant-depth models~\citep{Liu2022TransformersLS}.
Following \citet{Grazzi2024UnlockingSI}, we restrict each input to a swap of two elements, so that it can be rendered as a cup swap.
Every permutation of $k$ elements can be written as at most $k-1$ swaps~\citep{Grazzi2024UnlockingSI, Siems2025DeltaProductIS}.
For fixed $k$, this changes sequence length by only a constant factor, so the swap-only $S_5$ problem remains $\mathrm{NC}^1$-complete.

\subsection{Toy Experiments: State Tracking Without Pixels}
\label{sec:toy}
\begin{figure}[t]
    \centering
    \includegraphics[width=1.0\textwidth]{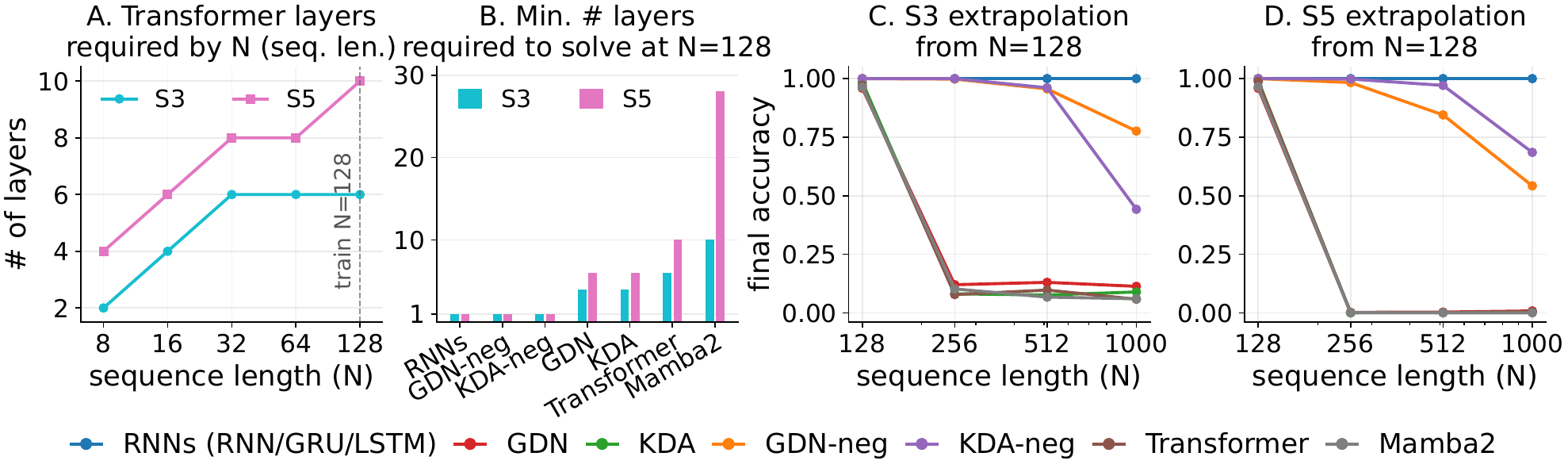}
    \vspace{-1.2em}
    \caption{
        \textbf{Pixel-free state tracking on the swap-only $S_3$ and $S_5$ word problems.}
        \textbf{(A)} Minimum number of Transformer layers required to solve a problem of sequence length $N$.
        \textbf{(B)} Minimum number of layers required by each architecture at $N{=}128$.
        \textbf{(C)} Length extrapolation on $S_3$ after training at $N{=}128$.
        \textbf{(D)} Length extrapolation on $S_5$ after training at $N{=}128$.
    }
    \vspace{-1.2em}
    \label{fig:toy-s3-s5-exp}
\end{figure}

We first run the synthetic state-tracking probe standard in this literature~\citep{Merrill2024TheIO, Grazzi2024UnlockingSI}, feeding $S_3$ and $S_5$ swap chains as raw symbol sequences.
To understand the linear-attention results, consider a DeltaNet-style layer, which carries a matrix state $S_t$ and updates it as
\begin{equation}
S_t = \big(I - \beta_t k_t k_t^\top\big) S_{t-1} + \beta_t k_t v_t^\top.
\label{eq:deltanet-update}
\end{equation}
The keys $k_t$ are unit-norm, so the transition $A_t = I - \beta_t k_t k_t^\top$ is a generalized Householder matrix with a single non-identity eigenvalue $1-\beta_t$.
The conventional range $\beta_t \in [0,1]$ confines this eigenvalue to $[0,1]$, where it can erase and overwrite but never reflect.
Widening to $\beta_t \in [0,2]$ reaches $[-1,1]$, and at $\beta_t = 2$ the transition becomes the exact reflection that realizes one swap~\citep{Grazzi2024UnlockingSI}.
As shown in Fig.~\ref{fig:toy-s3-s5-exp}, nonlinear RNNs apply a swap in a single layer, while linear-attention variants such as Gated DeltaNet (GDN,~\citet{Yang2024GatedDN}) and Kimi Delta Attention (KDA,~\citet{Zhang2025KimiLA}) do so only once their eigenvalues are allowed to span $[-1,1]$, which we mark with a ``-neg'' suffix throughout.
Mamba2 and Transformers need more layers as the target length grows and learn length-dependent shortcuts that do not extrapolate~\citep{Liu2022TransformersLS}.
Consistent with~\citet{Merrill2024TheIO}, even the easier, solvable group $S_3$ already forces the Transformer's depth to grow with sequence length (Fig.~\ref{fig:toy-s3-s5-exp}). 
We provide more experiment details in Appendix~\ref{app:toy}.

\subsection{The Visual Shell Game: Benchmark, Models, and Evaluation Protocol}

\begin{figure}[h]
    \centering
    \includegraphics[width=\textwidth]{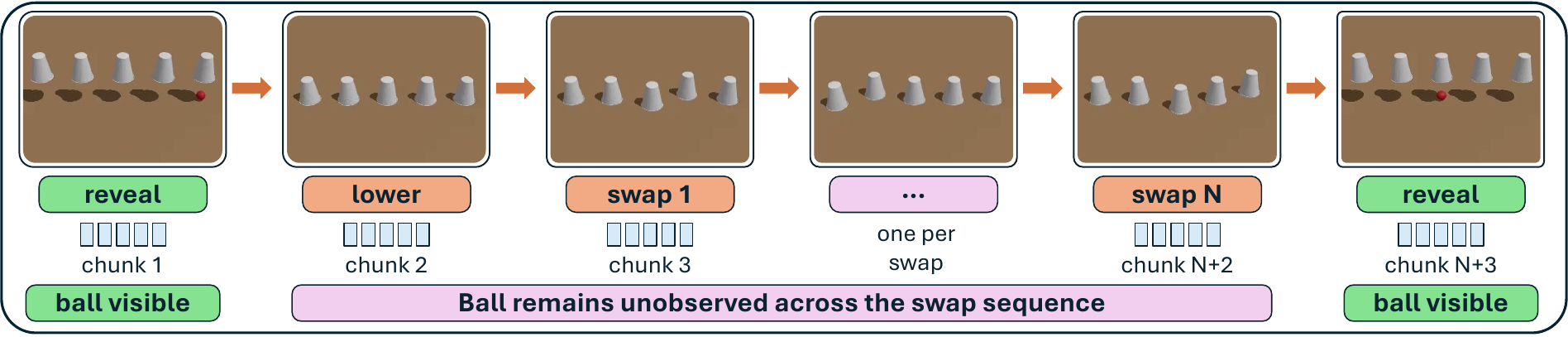}
    \vspace{-1.2em}
    \caption{
        \textbf{The visual shell game construction.}
        An episode first reveals the ball and lowers the cups, then applies a sequence of swaps while the ball remains hidden, and finally reveals its position.
    }
    \vspace{-0.5em}
    \label{fig:shellgame_construction}
\end{figure}

\label{sec:benchmarks-visual}
\noindent\textbf{Synthetic Data Generation.}
The Shell Game renders the swap-only $S_5$ task of Sec.~\ref{sec:formal} as action-conditioned video.\footnote{Unlike the full $S_5$ word problem or its swap-only variant above, which ask for the entire composed permutation, the visual Shell Game only asks for the ball's position. This is still a single coordinate of the same composed permutation, so producing it depends on the whole swap chain rather than any shortcut.}
We adapt the rendering code from \citet{Liu2026CanVM}, built on Three.js and WebGL, and add explicit lift and reveal phases together with action annotations (Fig.~\ref{fig:shellgame_construction}).
For $k$ cups, there are $\binom{k}{2}$ pairwise swap actions plus shared lift, lower, and no-op controls.
We use $k=5$ throughout and episodes are rendered at $256{\times}192$ resolution.
The cups are identical and the camera is fixed, so simply rendering their motion does not require knowing which cup holds the ball.
We align one swap with one autoregressive chunk of five latent frames, so each chunk advances the hidden permutation by one transposition.
An episode with $N$ swaps thus spans $N+3$ chunks in total, including two opening chunks and one final reveal chunk.
More details are in Appendix~\ref{app:visual-shell-game-dataset-generation}.

\begin{figure}[t]
    \centering
    \includegraphics[width=1.0\textwidth]{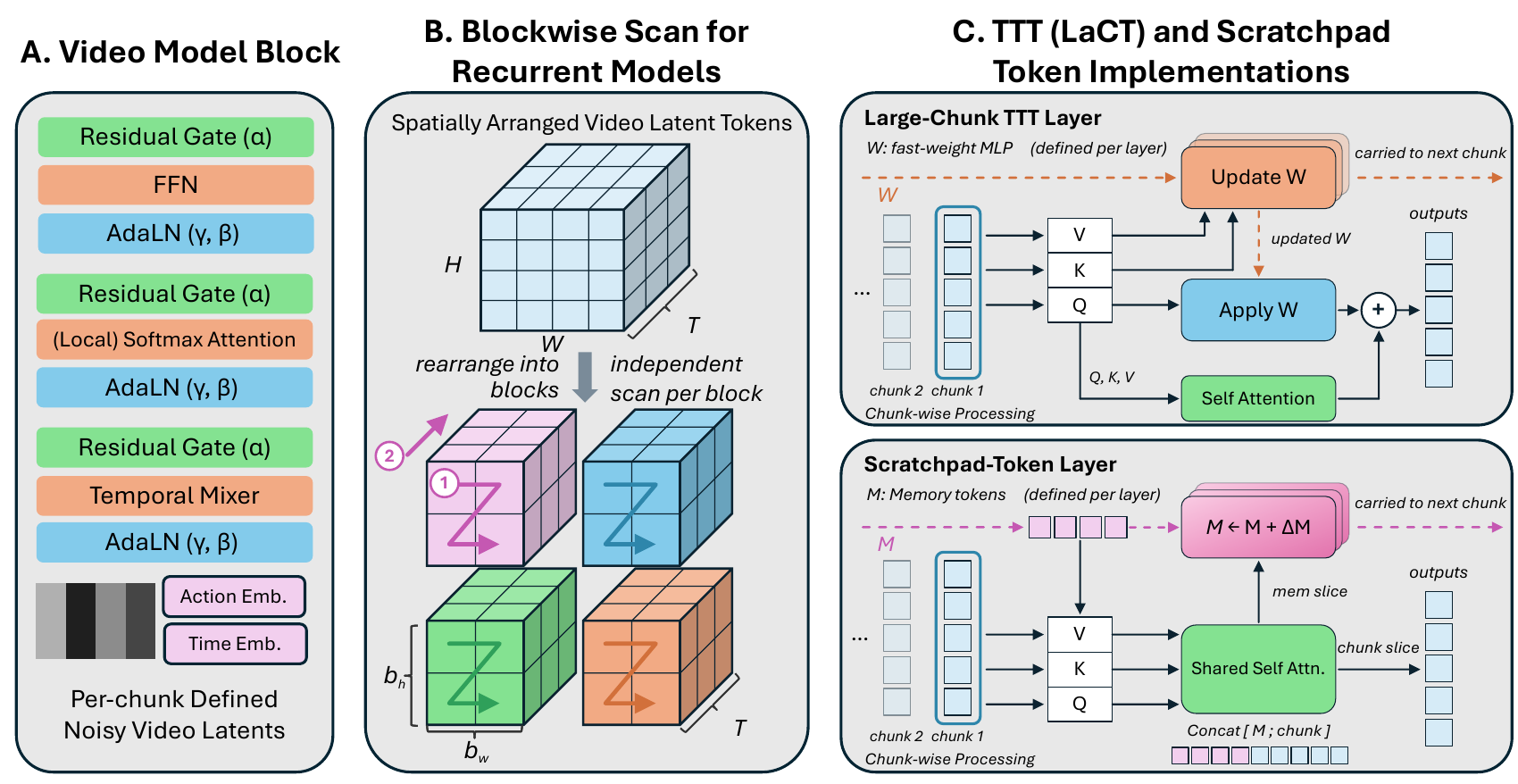}
    \vspace{-1em}
    \caption{\textbf{Architectural details.}
    \textbf{(A)} A canonical block stacks a temporal mixer, a sliding-window softmax attention with 3D RoPE, and an FFN as residual sublayers (pure-attention baselines drop the temporal mixer). Each sublayer applies standard AdaLN~\citep{Peebles2022ScalableDM}, with scale $\gamma$, shift $\beta$, and residual gate $\alpha$ regressed from the action and diffusion-timestep embeddings.
    \textbf{(B)} Following~\citet{Po2025LongContextSV}, we first reshape video tokens into an $H\times W\times T$ spatial latent grid and split into $b_h\times b_w\times T$ smaller blocks scanned independently, for better parallelism and a larger effective state.
    \textbf{(C)} Two examples of persistent per-layer recurrent states carried across autoregressive rollouts. LaCT updates fast-weight MLP $W$ from the chunk's keys/values, applies it to the queries, and sums with a self-attention branch. The scratchpad keeps $M$ memory tokens updated in place ($M\leftarrow M+\Delta M$) via shared self-attention over a concatenated sequence of $[M;\text{chunk}]$.}
    \vspace{-1em}
    \label{fig:model_architecture}
\end{figure}

\noindent\textbf{Model Architectures.}
All variants use a Wan-style latent video diffusion transformer with a frozen Wan VAE~\citep{Wang2025WanOA}, keeping the width fixed at 1152 and adjusting the depth to about 200M parameters.
The VAE compresses the video by $8\times8$ spatially and $4\times$ temporally.
We compare bidirectional and autoregressive Transformers~\citep{Chen2024DiffusionFN}, RNN/LSTM/GRU mixers, Mamba2~\citep{Dao2024TransformersAS}, GDN~\citep{Yang2024GatedDN}, KDA~\citep{Zhang2025KimiLA}, LaCT~\citep{Zhang2025TestTimeTD}, and explicit scratchpad tokens.
The recurrent variants combine a temporal mixer, sliding-window attention (SWA) with 3D RoPE, and a feedforward network as residual sublayers (Fig.~\ref{fig:model_architecture}), whereas pure-attention DiTs keep only the attention and feedforward sublayers.
RNNs and linear-attention variants use the spatial block-wise scan following \citet{Po2025LongContextSV} (Fig.~\ref{fig:model_architecture}B).
We write SWA$w$ for attention over the current and $w$ previous chunks and LaCT$H$ for LaCT with $H$ fast-weight heads, and use SWA1 and LaCT8 as the defaults throughout the paper.

\noindent\textbf{Training and Evaluation Protocol.}
All models train from scratch on 30,000 five-swap episodes for 200K steps with diffusion forcing and flow matching.
Each training episode spans eight chunks, or 157 decoded frames.
We use NorMuon~\citep{Li2025NorMuonMM} with a batch size of 16 and learning rate $2\times10^{-4}$ in mixed precision.
LaCT and the scratchpad update their carried state on each clean chunk before reading it to denoise the next one. Additional details on the model architecture, and training details such as clean/noisy packing and optimizer characteristics, are in Appendices~\ref{app:visual-shell-game-model-architecture} and~\ref{app:visual-shell-game-training-protocol}. For evaluation, we test from training length $N{=}5$ through $N{=}30$.\footnote{The causal full-attention DiT is trained on 8 chunks, so at longer rollouts we cap its attention to the current and 7 previous chunks, matching its training context. With unrestricted context, generation degrades on temporal RoPE offsets never seen in training. SWA variants are evaluated at their training window sizes, and we observe no visual-quality degradation. The bidirectional DiT admits no such capping and is evaluated with full context, which often degrades its outputs at longer lengths.}
Each length has $5\times10^N$ possible episodes from five start positions and ten choices per swap.
We score visual fidelity with PSNR, SSIM, and LPIPS, and state tracking with the final ball-position accuracy (chance $1/5$).
We evaluate 64 episodes at chain lengths up to 30 swaps and use 50 sampling steps (Appendix~\ref{app:visual-shell-game-evaluation-protocol}).

\section{Do Current Video Backbones Track Hidden State?}
\label{sec:failures}
\subsection{Standard Backbones Fail: Visual Fidelity Does Not Imply State Accuracy}
\label{sec:allfail}
\begin{figure}[t]
    \centering
    \includegraphics[width=1.0\textwidth]{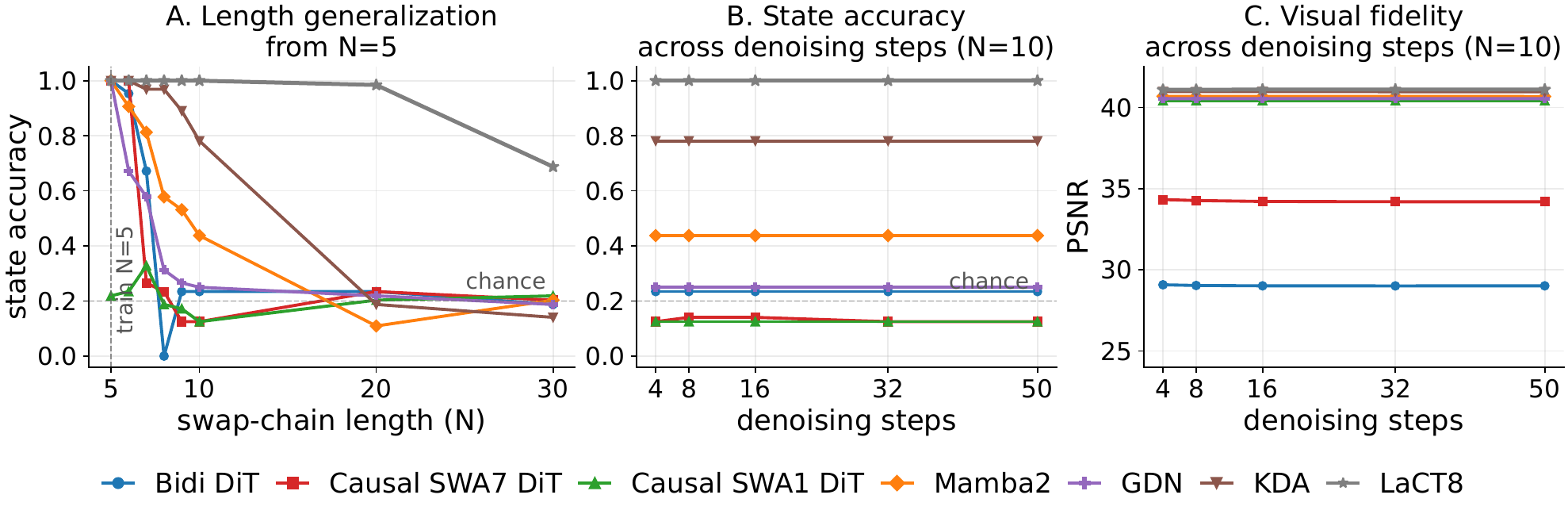}
    \vspace{-1.5em}
    \caption{
        \textbf{Shell Game state tracking and visual fidelity under length and denoising-step sweeps.} All recurrent and linear-attention variants share an SWA1 backbone; the causal DiTs are labeled by their window size. 
        \textbf{(A)} Length generalization for representative video backbones trained at $N{=}5$ swaps and evaluated through $N{=}30$: most standard backbones fit only short horizons and fall toward chance, while LaCT8 retains substantially higher state accuracy. 
        \textbf{(B)} At the extrapolation length $N{=}10$, increasing the denoising budget from $4$ to $50$ sampling steps does not improve state accuracy. 
        \textbf{(C)} The same $N{=}10$ denoising-step sweep shows high PSNR across models even when state accuracy is near chance, separating visual fidelity from hidden-state tracking.
    }
    \vspace{-1em}
    \label{fig:shellgame_main}
\end{figure}

\noindent\textbf{Standard backbones learn only shortcuts.}
Using the setup above, we first test the standard backbones commonly used today: bidirectional and autoregressive (AR) Transformers~\citep{Peebles2022ScalableDM, Yin2024FromSB}, Mamba2+SWA~\citep{Po2025LongContextSV,Savov2025StateSpaceDiffuserBL}, and nonnegative-eigenvalue linear-attention+SWA families~\citep{Li2026AttendLR, Zhu2026SANAWMEM}.
Trained for a fixed 200K steps, all full-context attention and recurrent variants except the SWA1 baselines reach 100\% accuracy at the in-distribution length, but none of them extrapolates robustly (Fig.~\ref{fig:shellgame_main}).
This pattern is consistent with the length-dependent shortcuts seen in the toy $S_5$ experiments (Sec.~\ref{sec:toy}).
Nonetheless, we find that visual fidelity stays high throughout the action sequence, with the model rendering plausible frames even when it fails to track the hidden state.
At evaluation we keep each softmax-attention variant within the temporal context it saw during training whenever the architecture allows, so that failures reflect state tracking rather than out-of-distribution RoPE offsets.

\noindent\textbf{Increasing denoising steps does not help.}
Consistent with the serial-depth analysis~\citep{Liu2025TheSS}, increasing diffusion sampling from 4 to 50 steps at $N=10$ does not change state accuracy (Fig.~\ref{fig:shellgame_main}).
LaCT8 already reaches $1.0$ accuracy at 4 steps, while the failing variants do not improve with more steps.
Visual fidelity also changes little because the Shell Game videos are very simple.

\subsection{Inspecting the Autoregressive Cache}
\label{sec:cache}
\noindent\textbf{Autoregressive KV cache is a visual history, not a compact running state.}
For autoregressive Transformers, the KV cache is the main channel that could carry the hidden arrangement across chunks.
To probe what it actually carries, we train AR Transformers with different SWA sizes.
\begin{wrapfigure}{r}{0.60\textwidth}
    \vspace{-0.5em}
    \centering
    \includegraphics[width=\linewidth]{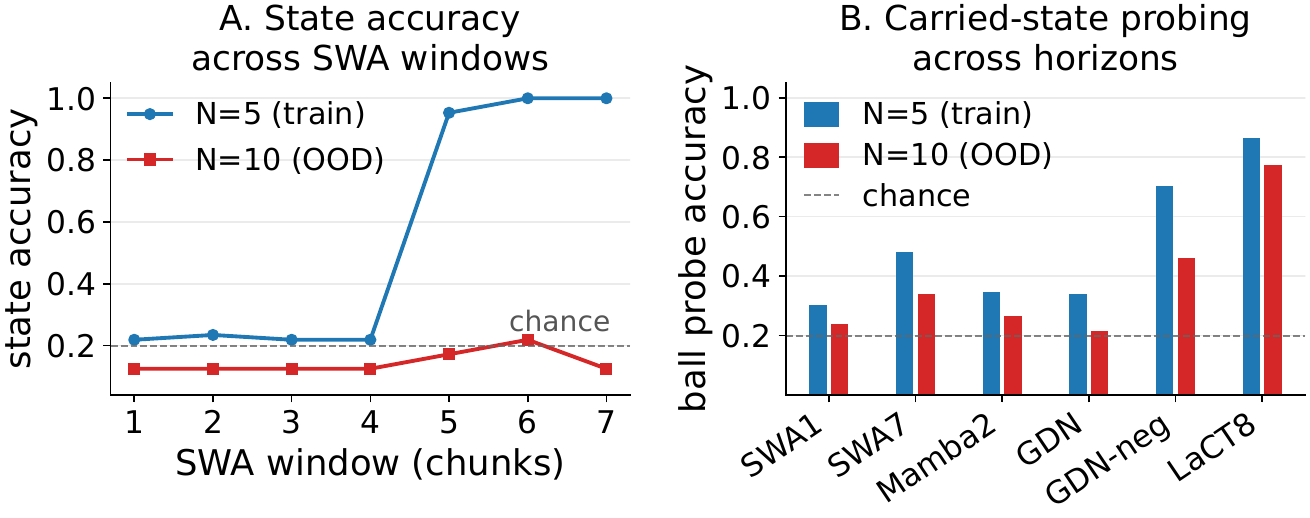}
    \vspace{-2em}
    \caption{
        \textbf{SWA size ablation and hidden-state decodability.}
        \textbf{(A)} Windows covering the full training swap history fit $N=5$, but all fail at $N=10$. SWA7 spans all eight chunks of a training episode.
        \textbf{(B)} Best-layer hidden-ball accuracy from $B=16$ fixed-bank reads over swap chunks at $N=5$ and $N=10$.
        GDN-neg and LaCT8 retain the strongest decodable state.
        Recurrent variants use SWA1.
    \vspace{-0.5em}
    \label{fig:shellgame_cache_probe}
}
\end{wrapfigure}
As shown in Fig.~\ref{fig:shellgame_cache_probe}, SWA1 through SWA4 fail to track the hidden state even at the training length of 5 swaps, and SWA5 is the first setting in which the final reveal can attend to all five preceding swap chunks.
SWA5 and wider windows fit the training length but still fail to extrapolate, including full-context attention.
A wider window exposes more history, but cached entries remain fixed once written.
A chunk can transform a retrieved state only at a higher layer, so composing swaps through the cache still consumes the model's fixed layer depth (Appendix~\ref{app:no-scratchpad}).
A Transformer language model could sidestep this by writing an intermediate result into an output token and reading it back at the next step.
However, this is impossible here, since our generated video chunks are trained only to reconstruct visible cup motion and have no target for the hidden ball.
All experiments use one diffusion-forcing pixel loss, with chunk-wise causal attention via FlexAttention~\citep{Dong2024FlexAA}.
Rollout training (with BPTT), a structured latent state, or task-specific rewards likely help, but we avoid such handcrafting here.

\noindent\textbf{Probing the internal states of autoregressive backbones.}
We analyze the internal states and caches of AR models, and since these are often in complex form (\textit{e.g.,} matrix states or SwiGLU fast weights), we do so by indirectly probing their readouts.
For each chunk $t$ and layer $\ell$, we save the model's own readout $y_t^\ell$ from the residual update $x_t^{\ell+1}=x_t^\ell+g_t^\ell y_t^\ell$, where $g_t^\ell$ is the gate.
$y_t^\ell$ is what the model passes forward to render the current chunk, and is a readout of the model's internal state given the input query.
We also save fixed-bank readouts of the carried state, probing it with a bank of $B$ fixed random Gaussian vectors ${p}_{m}$ shared across all chunks.
Depending on the backbone, $p_m$ serves as an attention query into the KV cache, $r_{t,m}^{\mathrm{KV}}=\mathrm{Attn}(p_m,K_{\leq t},V_{\leq t})$, as a linear read of a matrix recurrent state, $r_{t,m}^{S}=S_t^\top p_m$, or as an input to LaCT's fast weight, $r_{t,m}^{\mathrm{FW}}=f_{W_t}(p_m)$.
At 5 swaps (in-distribution) and 10 swaps (extrapolation), we train linear probes on $y_t^\ell$ and on the concatenated bank reads $\{r_{t,m}^\ell\}_{m=1}^B$, predicting the swap action or ball position.
The best-layer current-readout probes recover the swap action at $1.0$ accuracy for every model and horizon, but decode the hidden ball position only near chance in most cases (not shown in the paper).
The fixed-bank probes give a different result (Fig.~\ref{fig:shellgame_cache_probe}).
With $B=16$, ball accuracy at 5/10 swaps is $0.87/0.78$ for LaCT and $0.70/0.46$ for GDN-neg, noticeably higher than for the other backbones.
These results suggest that the failing AR backbones mostly carry action- and retrieval-bound content, while the successful recurrent mechanisms maintain a separate state variable for the hidden swap sequence.

\section{Mechanisms That Enable State Tracking}
\label{sec:mechanisms}
\begin{figure}
    \centering
    \includegraphics[width=\textwidth]{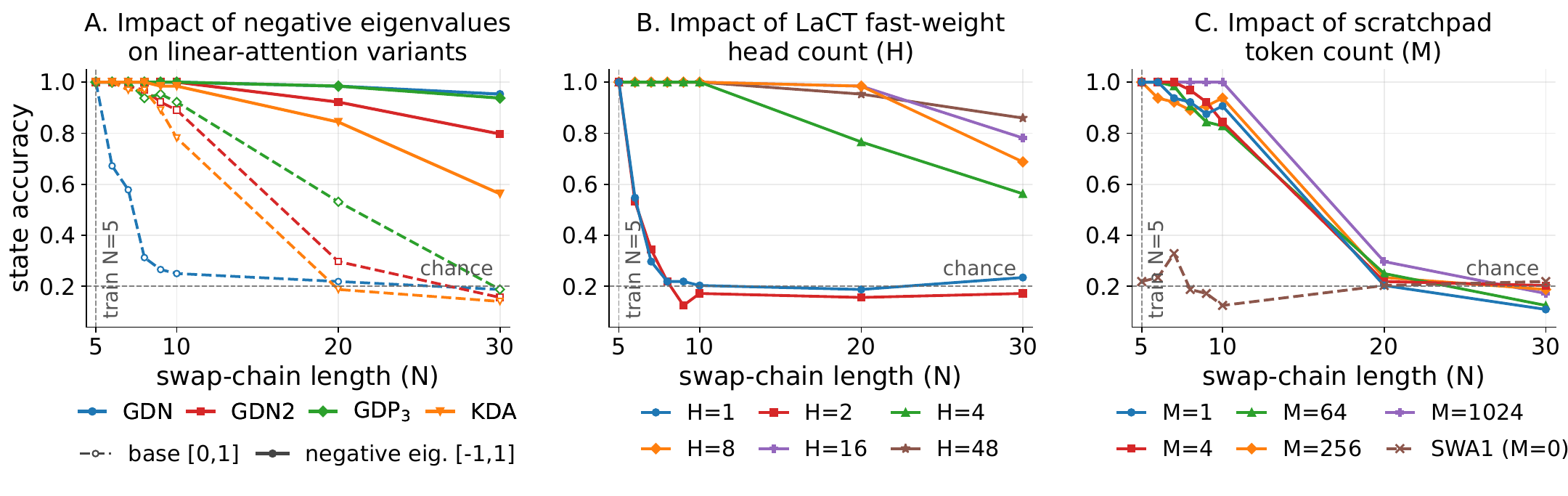}
    \vspace{-2em}
    \caption{
        \textbf{Length extrapolation of mechanisms that enable state tracking.} 
        All variants share the same SWA1 backbone, with the temporal mechanism added on top. The bare backbone alone (SWA1, $M{=}0$ in (C)) never rises above chance, failing even at the in-distribution length $N{=}5$, so it serves as the natural chance-level control.
        \textbf{(A)} Extending the transition eigenvalue range from $[0,1]$ to $[-1,1]$ improves state accuracy across linear-attention variants.
        \textbf{(B)} LaCT extrapolation improves with the number of fast-weight heads $H$.
        \textbf{(C)} Adding explicit per-layer read/writable scratchpad tokens improves extrapolation through $N{=}10$ but fails by $N{=}20$.
    }
    \label{fig:state_tracking_mechanisms}
    \vspace{-1em}
\end{figure}

From these failures, we hypothesize two key components that enable state tracking: \emph{a revisable state carried across chunks} paired with \emph{an update rule expressive enough to compose the hidden transition on that state}.
The AR Transformers lack a revisable state, and the nonnegative-eigenvalue linear attentions lack the expressive rule.
Here, we show two realizations of this pair, state transitions that allow negative eigenvalues (Sec.~\ref{sec:negeig}) and TTT-KVB with a nonlinear inner model (Sec.~\ref{sec:fastweight}), together with an explicit-token control that carries a state but only updates it additively (Sec.~\ref{sec:scratchpad}).

\subsection{Negative-Eigenvalue Linear Attention}
\label{sec:negeig}
Recalling the DeltaNet-style state update in Eq.~\ref{eq:deltanet-update} from Sec.~\ref{sec:toy}, widening $\beta_t\in[0,1]$ to $\beta_t\in[0,2]$ allows the transition to reach the full range $[-1,1]$.
We directly allow negative eigenvalues across our linear-attention variants, including up-to-date models such as Gated Delta Product (GDP,~\citet{Siems2025DeltaProductIS}) and Gated DeltaNet2 (GDN2,~\citet{Hatamizadeh2026GatedDD}), and find consistent improvements across all variants (Fig.~\ref{fig:state_tracking_mechanisms}), with GDN-neg essentially solving the task ($0.95$ at $N=30$).

\subsection{Nonlinear Fast Weights as a Recurrent State Update}
\label{sec:fastweight}
A second working mechanism comes from LaCT~\citep{Zhang2025TestTimeTD}, a TTT-KVB layer whose recurrent state is given by the parameters of a small inner network.
We use the bias-free SwiGLU
\begin{equation}
f_W(x)=W_1\!\left[\operatorname{silu}(W_0x)\odot(W_2x)\right],
\qquad W=(W_0,W_1,W_2).
\end{equation}
LaCT aggregates the gradient over each clean chunk $\mathcal{C}_t$ and applies one update to the fast weights, which are then read by the following noisy chunk.
We use the default negative dot-product KV-binding objective, and ablate it against an MSE objective,
\begin{equation}
\ell_t^{\mathrm{dot}}
    =-\sum_{i\in\mathcal{C}_t}\left\langle f_W(k_i),v_i\right\rangle,
\qquad
\ell_t^{\mathrm{mse}}
    =\frac{1}{2}\sum_{i\in\mathcal{C}_t}
      \left\lVert f_W(k_i)-v_i\right\rVert_2^2.
\end{equation}
The matrices $W_0,W_1,W_2$ persist across autoregressive chunks and are updated inside each layer, giving the model an explicit recurrent state that the video objective can learn to use.

\noindent\textbf{More fast-weight heads improve tracking despite smaller raw state size.}
Fig.~\ref{fig:state_tracking_mechanisms} shows that one or two heads fail beyond the training length, whereas four or more extrapolate.
\begin{wraptable}{r}{0.55\textwidth}
    \centering
    \vspace{-0.7em}
    \small
    \caption{
        \textbf{LaCT8 fast-weight ablations on the Shell Game.} 
        We report state accuracy at swap-chain length $N$. All variants use eight fast-weight heads and one online update per chunk. ``Freeze'' denotes parameters excluded from the online fast-weight update.
    }
    \label{tab:lact-ablation}
    \begin{tabular}{lcccc}
        \toprule
        Variant & $N{=}5$ & $N{=}10$ & $N{=}20$ & $N{=}30$ \\
        \midrule
        LaCT8 baseline & 1.00 & 1.00 & 0.98 & 0.69 \\
        + 1-layer MLP & 1.00 & 0.23 & 0.23 & 0.13 \\
        + Freeze $W_0,W_2$ & 1.00 & 0.14 & 0.30 & 0.16 \\
        + Freeze $W_1$ & 1.00 & 1.00 & 0.97 & 0.64 \\
        + Gradient ascent & 1.00 & 1.00 & 0.95 & 0.72 \\
        + MSE loss & 0.22 & 0.16 & 0.22 & 0.22 \\
        \bottomrule
    \end{tabular}
    \vspace{-1em}
\end{wraptable}
At fixed model width $D$, each head has width $d=D/H$.
With SwiGLU expansion ratio $r$, $W_0,W_2\in\mathbb{R}^{rd\times d}$ and $W_1\in\mathbb{R}^{d\times rd}$, so the fast weights carry $H(3rd^2)=3rD^2/H$ scalars per layer (Appendix~\ref{app:visual-shell-game-model-architecture}).
Increasing $H$ therefore reduces the state size while splitting the update across more independently controlled heads, and the latter appears to matter more for state tracking in this experiment.
With a single head, the same fast-weight state must support both rendering and state updates, whereas multiple heads may allow individual channels to specialize in task-relevant state.

\noindent\textbf{A single-layer linear inner model is insufficient.}
Replacing LaCT's SwiGLU with a single-layer MLP preserves recurrent storage and fits the training length, but fails to extrapolate (Tab.~\ref{tab:lact-ablation}).
For a linear inner map $f_S(x)=S^\top x$, a dot-product gradient step gives $S_t=S_{t-1}+\eta k_t v_t^\top$, the additive outer-product update of plain linear attention (Appendix~\ref{app:ttt-linear-attention}).
By contrast, the SwiGLU inner model produces a history-dependent feature map that does not admit the same closed-form associative reduction, giving LaCT a richer update class than plain linear attention~\citep{Liu2026TestTimeTW}.

\noindent\textbf{State tracking depends on the online feature-map update.}
Re-writing the SwiGLU inner model,
\begin{equation}
\phi_t(x)=\operatorname{silu}(W_{0,t}x)\odot(W_{2,t}x),
\qquad
f_{W_t}(x)=W_{1,t}\phi_t(x),
\end{equation}
separates the evolving feature map $(W_0,W_2)$ from the final readout matrix $W_1$.
As shown in Tab.~\ref{tab:lact-ablation}, updating $W_1$ alone fails, whereas updating $W_0$ and $W_2$ with $W_1$ frozen still retains most of the full update.
This shows that state tracking relies on the evolving $(W_0,W_2)$ feature map and the history-dependent kernel it induces, rather than on updates to $W_1$ alone.
By changing the kernel with each swap rather than processing every swap in a fixed feature space, the autoregressive Transformers with fast weights can implement the compositional updates required for $\mathrm{NC}^1$-hard state tracking.

\noindent\textbf{The inner loss does not determine the tracking mechanism.}
As shown in Tab.~\ref{tab:lact-ablation}, reversing the sign of the dot-product update leaves tracking largely unchanged.
Consistent with the gradient-ascent observation of \citet{Liu2026TestTimeTW}, end-to-end training can absorb the sign of the fast-weight write.
We find the MSE inner loss more sensitive.
In the token-wise, single-layer linear case, MSE recovers the DeltaNet rule, yet our chunked SwiGLU model with an MSE inner loss fails even at the training length.
For a linear inner model, a chunked MSE update introduces a multiplicative transition whose scale can grow with chunk size, making the inner learning rate harder to tune (Appendix~\ref{app:ttt-linear-attention}).
The two losses also guide the write differently: the dot loss rewards movement in the value direction $v_t$, whereas MSE follows the full residual $v_t-f_W(k_t)$ toward exact reconstruction.
We tentatively hypothesize that, together with weight normalization, this simpler dot update gives LaCT a stable, well-regularized write space that outer training can organize around the hidden state, whereas MSE ties the same space more closely to visible, render-bound values.

\subsection{Explicit Scratchpad Tokens}
\label{sec:scratchpad}
The carried state does not have to live in the weights.
We also test a token-space register by adding $M$ per-layer memory tokens, which each block reads and writes through attention and carries additively across chunks as $M\leftarrow M+\Delta M$ (Fig.~\ref{fig:model_architecture}).
Adding a single memory token per layer (next to the $960$ video tokens of each chunk) already lifts the at-chance SWA1 baseline to perfect in-distribution accuracy and generalizes up to $N{=}10$ swaps.
However, accuracy drops to chance by $N{=}20$ swaps (Fig.~\ref{fig:state_tracking_mechanisms}) across all token counts from 1 to 1024.
We find that the plain additive update lets the carried memory norm explode on longer rollouts, while LaCT better normalizes states via fast weights and keeps extrapolating as a better-regularized register.

\section{Beyond the Canonical Shell Game}
\label{sec:beyond}
Our Shell Game experiments show that current AR video backbones cannot compose unobserved state updates, so they should fail whenever a correct frame depends on a long swap chain among five or more hidden objects.
General video world models are much harder.
Many of their prediction problems sit inside $\mathrm{TC}^0$ yet remain hard to learn, while some others plausibly require $\mathrm{NC}^1$-hard state tracking and hence lie outside $\mathrm{TC}^0$.
More importantly, the world state is rarely a pure function of the action stream: it must also be updated from observations, either clean context frames or the model's own generations, and parts of it dynamically keep evolving out of sight, driven by no action at all.
We close with an early exploration of two hard setups, which originally motivated this project.

Memory Maze~\citep{Paukonis2022EvaluatingLM, Po2025LongContextSV, Chen2025RecurrentAD} gives the model a clean exploration prefix and asks it to generate the frames that follow.
We compare providing the camera pose at every step with local-action conditioning, where the model must track its pose from the initial pose and discrete controls before retrieving the corresponding visual memory.
On the original benchmark, all our backbones lose fidelity with local actions, where the MuJoCo dynamics can produce different displacements for the same action (Appendix~\ref{app:maze}). 
We rebuild the dataset with constant-displacement kinematics so that, without collisions, the pose is a deterministic function of the initial pose and actions.
Tracking it is then an iterated addition of rotated unit steps in a virtually abelian subgroup of $\mathrm{SE}(2)$, which plausibly lies in $\mathrm{TC}^0$~\citep{Merrill2022ThePT}, unlike $S_5$ composition.
The local action/global pose conditioning gap largely closes in this setting (without collision).
Meanwhile, with wall collisions enabled, a blocked move becomes a no-op, so the update depends on the unobserved layout and must be corrected from what the model sees.
The textured 3D Block World~\citep{Lillemark2026FlowEW} tests the other case, where the blocks keep moving while the agent's camera is turned away, so parts of the world evolve with no action at all (Appendix~\ref{app:blockworld}).

Every backbone performs well under global pose conditioning but most of them fail when switched to local-action conditioning, indicating that the bottleneck is state (pose) tracking rather than pure retrieval. 
In most settings LaCT is one of the strongest backbones, whereas the negative-eigenvalue extension, which is decisive on the Shell Game, gives no consistent gain (Tabs.~\ref{tab:maze-collision} and~\ref{tab:blockworld-tex-backbones}).
We suspect that the algebraic structure of an action-only permutation chain does not directly carry over, while the nonlinear fast-weight update still helps somewhat.
We do not yet know which architectures can learn these updates.
Future models have to use observations to correct their state and recognize failed actions, while also predicting how unseen objects continue to evolve when no action arrives.

\section{Conclusion}
We studied whether video generative models can maintain a world state that is not directly visible, using a five-cup Shell Game as a visual analogue of swap-only $S_5$ state tracking.
Since the pixel-centric objective does not supervise the invisible state, the model must hold it implicitly, and standard backbones fail to do so beyond the training horizon.
Reliable tracking requires a revisable state carried across chunks, an update expressive enough to apply a swap, and a gradient path from each write to its later use, which linear attention with negative transition eigenvalues and Transformers with nonlinear fast weights both provide.
We further discuss cases where the world state is not determined by the action stream alone.
Defining and learning a state that captures both the visible and hidden parts of the world remains an open challenge for video world models, and we believe this will require progress in tokenization, architectures, and objectives that reach beyond visible content.

\bibliography{main}
\bibliographystyle{iclr2027_conference}

\clearpage
\appendix
{\centering\LARGE Appendix\par}
\setcounter{table}{0}
\setcounter{figure}{0}
\setcounter{equation}{0}
\renewcommand{\thefigure}{A\arabic{figure}}
\renewcommand{\thetable}{A\arabic{table}}
\renewcommand{\theequation}{A\arabic{equation}}
\section{Autoregressive Rollout Alone Does Not Train State Updates}
\label{app:no-scratchpad}

\begin{wrapfigure}{r}{0.52\textwidth}
    \vspace{-0.8em}
    \centering
    \includegraphics[width=\linewidth]{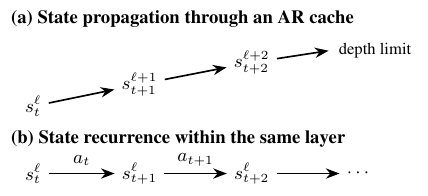}
    \caption{\textbf{State-update paths across autoregressive chunks.}
    The upper path moves to a higher layer after each update, while the lower path applies successive actions within one layer.
    $s_t^\ell$ denotes state at chunk $t$ and layer $\ell$, and $a_t$ is an action.}
    \label{fig:state-update-paths}
    \vspace{-1.2em}
\end{wrapfigure}

An autoregressive DiT passes cached features and generated chunks to the next model call.
Fig.~\ref{fig:state-update-paths} compares the cache path with a recurrent update within one layer.

\noindent\textbf{Cached features.}
A KV entry at layer $\ell$ is computed from that layer's input, so reading and updating it can only write the new state at a higher layer, where it becomes readable.
In the Shell Game, the initial reveal places the ball position in the cache at every layer.
For a ball position represented by $s_t^\ell$, a cache read followed by a swap update produces $s_{t+1}^{\ell+1}$, and repeating this at the next chunk produces $s_{t+2}^{\ell+2}$.
Each update along this path uses another layer, limiting successive updates to the model's depth (Fig.~\ref{fig:state-update-paths}).
In RNNs, linear attention, and TTT, each chunk instead updates the previous state in place at the same layer, so a single layer can simulate the whole swap chain as long as its update can express a swap.

\noindent\textbf{Generated chunks.}
At inference, a generated chunk returns as input and could in principle hold the ball position, as a CoT token holds an intermediate result~\citep{Merrill2024TheEP}.
Our training instead uses noisy ground-truth chunks encoded by the frozen Wan VAE~\citep{Wang2025WanOA}, and during a swap they show the cup motion but contain no label for the hidden ball.
In teacher/diffusion forcing, the next chunk receives the (noisy) ground-truth latent~\citep{Chen2024DiffusionFN}.
A mistake at the final reveal therefore has no gradient path back through the earlier generated chunks and cannot tell them what to preserve for that later prediction.
Repeated denoising does not change this training path.
Its calls share the same visual target, and we use a fixed number of calls per chunk as swap chains get longer.
More denoising loops did not improve tracking in our sweep, consistent with the analysis of \citet{Liu2025TheSS}.

Self Forcing uses the model's own generated chunks during training~\citep{Huang2025SelfFB}, but its practical implementation backpropagates through one selected denoising call per chunk and detaches the earlier cached representations.
A later loss still cannot train an earlier chunk to preserve information for the rest of the rollout.
Full rollout backpropagation, explicit state targets, or task-specific rewards could provide this supervision.
We have not tested these alternatives because of their computational cost or task-specific design, keeping the same video objective across our experiments.
Our recurrent variants backpropagate through the internal state's write and later use.

\section{Linear TTT Updates as Linear Attention and DeltaNet}
\label{app:ttt-linear-attention}

With a bias-free linear inner model, the TTT update can be written directly in terms of its keys and values.
We derive the additive linear-attention update from the dot-product loss and the DeltaNet update from MSE~\citep{Sun2024LearningT,Liu2026TestTimeTW}.

\paragraph{Setup.}
Let $S_t\in\mathbb{R}^{d_k\times d_v}$ be the fast weights of a bias-free linear inner model $f_{S_t}(k)=S_t^\top k$.
For each token $x_t$, $t=1,2,\ldots$, the slow (outer-loop) weights produce a query and key $q_t,k_t\in\mathbb{R}^{d_k}$ and a value $v_t\in\mathbb{R}^{d_v}$.
We take one gradient step on the inner loss $\ell_t$ at the current key--value pair,
$S_t=S_{t-1}-\eta_t\nabla_S\ell_t(S_{t-1})$,
and read the updated state with the query to obtain $o_t=S_t^\top q_t\in\mathbb{R}^{d_v}$.

\paragraph{Plain linear attention from the dot-product KV-binding loss.}
For the negative dot-product KV-binding objective $\ell_t(S)=-\langle S^\top k_t,v_t\rangle=-k_t^\top Sv_t$, the gradient can be calculated as $\nabla_S\ell_t=-k_tv_t^\top$.
Taking a single update step yields
\begin{equation}
S_t=S_{t-1}+\eta_t k_tv_t^\top,
\qquad
o_t=S_t^\top q_t=S_0^\top q_t+\sum_{i=1}^{t}\eta_i v_i(k_i^\top q_t),
\end{equation}
where $S_0$ is the state before the first update.
With $S_0=0$, the readout sums the values seen so far, weighted by their key--query inner products and learning rates, as in causal linear attention~\citep{Schlag2021LinearTA}.
This is the linear TTT-KVB equivalence described by \citet{Liu2026TestTimeTW}; summing the per-step updates also gives the batch gradient-descent form of \citet[Thm.~1]{Sun2024LearningT}.

\paragraph{DeltaNet from the MSE KV-binding loss.}
Now take the MSE objective $\ell_t(S)=\tfrac{1}{2}\lVert S^\top k_t-v_t\rVert^2$, whose gradient is $\nabla_S\ell_t=k_t(S^\top k_t-v_t)^\top$.
Taking one step here gives
\begin{equation}
S_t=S_{t-1}-\eta_t k_t(S_{t-1}^\top k_t-v_t)^\top
   =(I-\eta_t k_tk_t^\top)S_{t-1}+\eta_t k_tv_t^\top.
\end{equation}
Evaluating the updated state at the current key makes the overwrite explicit.
For $\lVert k_t\rVert_2=1$, $S_t^\top k_t=(1-\eta_t)S_{t-1}^\top k_t+\eta_t v_t$, so $\eta_t$ interpolates between the previous value associated with $k_t$ and the new value $v_t$.
At $\eta_t=1$, the update replaces the previous value exactly.
This is the DeltaNet delta rule with write strength $\beta_t=\eta_t$~\citep{Yang2024ParallelizingLT}, so online TTT-Linear with the MSE loss recovers DeltaNet~\citep{Sun2024LearningT}.

\paragraph{Chunked updates and learning-rate scaling.}
For a chunk $\mathcal{C}_m$ of size $c$, collect the keys and values as
$K_m=[k_i]_{i\in\mathcal{C}_m}$ and $V_m=[v_i]_{i\in\mathcal{C}_m}$, and let $D_m=\operatorname{diag}(\eta_i)\succeq0$ contain their learning rates.
A single gradient step on the chunk's summed loss, with all per-token gradients taken at the pre-update state $S_{m-1}$ gives
\begin{equation}
S_m^{\mathrm{dot}}=S_{m-1}+K_mD_mV_m^\top,
\qquad
S_m^{\mathrm{mse}}=(I-K_mD_mK_m^\top)S_{m-1}+K_mD_mV_m^\top.
\label{eq:chunked-dot-mse}
\end{equation}
For $c=1$, the MSE rule is exactly DeltaNet.
For $c>1$, it is a single batched-gradient step rather than a product of sequential DeltaNet updates.
Writing $G_m=K_mD_mK_m^\top$, the raw MSE transition is non-expansive in Frobenius norm exactly when the largest eigenvalue satisfies $\lambda_{\max}(G_m)\leq2$.
Larger chunks add positive-semidefinite rank-one terms to $G_m$, so its largest eigenvalue can grow, particularly with correlated keys.
For example, in our setting a single update aggregates the $c=960$ tokens of a video chunk, and without rescaling, a plain $\eta=10^{-2}$ already gives $\lambda_{\max}(G_m)=\eta c=9.6$ in this worst case, far above the non-expansive bound of $2$.
The dot rule has no corresponding multiplicative transition in this linear setting.

For the SwiGLU inner model of Sec.~\ref{sec:fastweight}, the update of $W_1$ has the same form with $K_m$ replaced by $\Phi_m=[\phi(k_i)]_{i\in\mathcal{C}_m}$, where $\phi(x)=\operatorname{silu}(W_0x)\odot(W_2x)$.
Freezing $\phi$ leaves chunked linear attention with a fixed kernel.
This variant fails past the training length in Tab.~\ref{tab:lact-ablation}.
Full LaCT also updates $W_0$ and $W_2$ and restores the row norms of all three fast matrices after each chunk.

\paragraph{Connection to the Shell Game inner-loss ablation.}
In our chunked SwiGLU ablation, the dot-product objective preserves state tracking, whereas MSE fails even at the training horizon (Tab.~\ref{tab:lact-ablation}).
We hypothesize that the dot-product objective works better in our setting because it provides a less restrictive binding signal, rewarding a larger inner product between $f_W(k)$ and $v$ without requiring an exact match.
MSE instead drives $f_W(k)$ toward exact reconstruction of the projected visual value $v$, which we believe leaves less freedom for the fast weights to encode unobserved state not specified by the current visual target.
As shown above for a simple linear inner model, the MSE residual also introduces a state-dependent multiplicative transition that can become unstable as the chunk grows.
Even under the dot-product objective, full LaCT does not reduce to a static-kernel prefix sum, since updating $W_0$ and $W_2$ online makes its nonlinear feature map history-dependent~\citep{Liu2026TestTimeTW}.

\section{Toy State-Tracking Experiment Details}
\label{app:toy}

\paragraph{Task.}
We use the symmetric-group word problems $S_3$ and $S_5$ in the swap-only setting~\citep{Grazzi2024UnlockingSI, Siems2025DeltaProductIS}. 
Each action is a transposition of two elements, and the target at every step is the running prefix product of the actions so far.
The model receives a single initial state and the action stream and must predict the current permutation at each step.

\paragraph{Models and Depth Sweep.}
We compare nonlinear RNNs (RNN, GRU, LSTM), a Transformer, Mamba2, Gated DeltaNet (GDN), and Kimi Delta Attention (KDA), together with GDN-neg and KDA-neg, which widen the transition spectrum from $[0,1]$ to $[-1,1]$.
The RNN, GRU, and LSTM initialize their recurrent state from the initial permutation, while the remaining backbones prepend it as an input token.
We keep width fixed within each family during the depth sweep, choosing widths to roughly match parameters per layer.
These are $256$ for the Transformer, $288$ for Mamba2, $304$ for GDN/KDA, and $644/376/324$ for RNN/GRU/LSTM.
The Transformer uses four attention heads, Mamba2 uses a state dimension of $16$, a convolution width of $4$, and an expansion ratio of $2$, and GDN and KDA use four recurrent heads.

All models are trained from scratch on sequences of $N=128$ swaps for up to $300$K steps using AdamW with a batch size of $256$ and a learning rate of $10^{-4}$.
We count a length as solved when the best validation checkpoint reaches $0.95$ final-step accuracy.
Chance is $1/6$ on $S_3$ and $1/120$ on $S_5$.
Fig.~\ref{fig:toy-s3-s5-exp} first evaluates the Transformer depth sweep at lengths $N\in\{8,16,32,64,128\}$ and reports the minimum depth that solves each length.
It then applies the same criterion across architectures at $N=128$, starting from one layer and increasing over the tested even depths until the threshold is reached.
For the extrapolation results, we take the minimum-depth model that solves $N=128$ for each architecture and evaluate it at $N\in\{128,256,512,1000\}$.
Since the RNN, GRU, and LSTM all solve both tasks with a single layer, we show the vanilla RNN as their representative in the figure.

\section{The Visual Shell Game Experiment Details}
\label{app:visual-shell-game}

\label{app:visual-shell-game-dataset-generation}

We adapt the Three.js and WebGL cups-and-ball renderer of \citet{Liu2026CanVM} to render videos at $256{\times}192$, adding explicit lift and reveal phases together with action annotations.
The action vocabulary contains $\binom{k}{2}$ pairwise swaps and three global controls for lifting all cups, lowering them, and no-op leaving the scene unchanged, giving ten swaps and thirteen actions in total for five cups.

We divide each episode into an initial reveal, lowering, $N$ swaps, and a final reveal, assigning one chunk of five latent frames to each reveal, lowering phase, and swap.
Only the opening and closing reveals show the ball.
An episode has $N+3$ chunks, or $5(N+3)$ latent frames, and the Wan 2.1 VAE decodes the first latent to one pixel frame and each subsequent latent to four more, giving
\[
4\cdot5(N+3)-3 = 20N+57
\]
pixel frames.
We train on five swaps, with 8 chunks, 40 latents, and 157 pixel frames, and test through 30 swaps, with 33 chunks, 165 latents, and 657 pixel frames.
For any length $N$, five start positions and ten choices at each swap give $5\times10^N$ possible episodes.
This combinatorial growth makes memorization impossible, so we use length extrapolation as the primary diagnostic.

\subsection{Model Architecture}
\label{app:visual-shell-game-model-architecture}

All variants share a Wan-style latent video diffusion scaffold~\citep{Wang2025WanOA}.
The frozen Wan2.1 VAE maps each $256{\times}192$ video to 16-channel latents with $4{\times}$ temporal and $8{\times}$ spatial compression.
Each autoregressive chunk contains five latent frames on a $32{\times}24$ grid, and $1{\times}2{\times}2$ patches produce $960$ tokens per chunk.

We set the model width to $D=1152$ with 16 attention heads of dimension 72, QK normalization, 3D RoPE, and a two-layer GELU FFN of width $4D$.
For conditioning, we project a 256-dimensional sinusoidal timestep embedding and a 72-dimensional lookup for the 13 actions to $D$ and inject both through AdaLN.
The pure-attention and scratchpad models use 12 layers, while the recurrent families adjust their depth to remain at a broadly comparable scale around $200$M parameters.
Causal models use SWA1 by default, attending to the current and previous chunk.

Tab.~\ref{tab:shellgame-backbone-architecture} reports the layer count, internal mixer geometry, recurrent transition range, and exact parameter count for the standard and linear-attention backbones.
RNN, Mamba2, and linear-attention models have eight layers, with spatial block sizes $[1,1,1,1,2,2,4,4]$ from bottom to top.
The linear-attention variants use 12 recurrent heads of width 96.
Their negative-eigenvalue counterparts keep the same dimensions and change only the transition parameterization from $[0,1]$ to $[-1,1]$.

\begin{table}[ht]
\centering
\caption{\textbf{Shell Game backbone configurations.}
SWA$w$ attends to the current and $w$ previous chunks, the ``-neg'' suffix widens the transition spectrum from $[0,1]$ to $[-1,1]$, and GDP3 denotes Gated Delta Product with three Householder factors.
$L$ denotes depth and Params denotes the parameter count in millions.}
\label{tab:shellgame-backbone-architecture}
\scriptsize
\setlength{\tabcolsep}{3.5pt}
\begin{tabular*}{\textwidth}{@{\extracolsep{\fill}}lclcc}
\toprule
Backbone & $L$ & Mixer configuration & Transition range & Params (M) \\
\midrule
Bidirectional DiT & 12 & None with full-clip attention & -- & 201.503 \\
Causal DiT (training context, SWA7) & 12 & None with SWA7 attention & -- & 201.503 \\
Causal DiT (SWA1) & 12 & None with SWA1 attention & -- & 201.503 \\
RNN + SWA1 & 8 & Vanilla RNN with width $1152$ & nonlinear & 162.988 \\
GRU + SWA1 & 8 & GRU with width $1152$ & nonlinear & 205.492 \\
LSTM + SWA1 & 8 & LSTM with width $1152$ & nonlinear & 226.744 \\
Mamba2 + SWA1 & 8 & $d_{\mathrm{state}}=256$, $d_{\mathrm{conv}}=4$, expand $2$, head dim. $72$ & positive & 210.582 \\
GDN + SWA1 & 8 & $12$ heads of width $96$ & $[0,1]$ & 195.153 \\
GDN-neg + SWA1 & 8 & $12$ heads of width $96$ & $[-1,1]$ & 195.153 \\
GDN2 + SWA1 & 8 & $12$ heads of width $96$ & $[0,1]$ & 209.106 \\
GDN2-neg + SWA1 & 8 & $12$ heads of width $96$ & $[-1,1]$ & 209.106 \\
GDP3 + SWA1 & 8 & $12$ heads of width $96$ with $3$ Householder factors & $[0,1]$ & 237.989 \\
GDP3-neg + SWA1 & 8 & $12$ heads of width $96$ with $3$ Householder factors & $[-1,1]$ & 237.989 \\
KDA + SWA1 & 8 & $12$ heads of width $96$ & $[0,1]$ & 187.983 \\
KDA-neg + SWA1 & 8 & $12$ heads of width $96$ & $[-1,1]$ & 187.983 \\
\bottomrule
\end{tabular*}
\end{table}

LaCT and the explicit scratchpad carry persistent per-layer state across chunks, with their configurations summarized in Tab.~\ref{tab:shellgame-register-capacity}.
For LaCT we set the SwiGLU expansion to $r=2$.
With $H$ heads, each has width $d=D/H$ and three fast matrices containing $3rd^2$ scalars.
The total state per layer is therefore $3rD^2/H$.
All LaCT variants use a dot-product binding loss (except for the MSE ablations), a base inner-loop learning rate of $10^{-3}$, row-wise weight normalization, and one fast-weight update per clean chunk.
The scratchpad carries $M$ vectors of width $D$ in each layer and updates them additively once per chunk.

\begin{table}[ht]
\centering
\caption{\textbf{LaCT and scratchpad state configurations.}
$L$ denotes depth and Params reports the total model parameter count in millions.}
\label{tab:shellgame-register-capacity}
\scriptsize
\setlength{\tabcolsep}{4pt}
\begin{tabular*}{\textwidth}{@{\extracolsep{\fill}}lcllrr}
\toprule
Variant & $L$ & Heads / tokens & State structure per layer & State Size / layer & Params (M) \\
\midrule
LaCT1 & 8 & $H=1$ & $d=1152$, three SwiGLU matrices & 7,962,624 & 201.507 \\
LaCT2 & 10 & $H=2$ & $d=576$, three SwiGLU matrices & 3,981,312 & 209.559 \\
LaCT4 & 11 & $H=4$ & $d=288$, three SwiGLU matrices & 1,990,656 & 207.674 \\
LaCT8 & 12 & $H=8$ & $d=144$, three SwiGLU matrices & 995,328 & 213.850 \\
LaCT16 & 12 & $H=16$ & $d=72$, three SwiGLU matrices & 497,664 & 208.209 \\
LaCT48 & 12 & $H=48$ & $d=24$, three SwiGLU matrices & 165,888 & 205.555 \\
\midrule
Scratchpad1 & 12 & $M=1$ & $1{\times}1152$ additive memory & 1,152 & 201.517 \\
Scratchpad4 & 12 & $M=4$ & $4{\times}1152$ additive memory & 4,608 & 201.558 \\
Scratchpad64 & 12 & $M=64$ & $64{\times}1152$ additive memory & 73,728 & 202.388 \\
Scratchpad256 & 12 & $M=256$ & $256{\times}1152$ additive memory & 294,912 & 205.042 \\
Scratchpad1024 & 12 & $M=1024$ & $1024{\times}1152$ additive memory & 1,179,648 & 215.659 \\
\bottomrule
\end{tabular*}
\end{table}

\subsection{Training Protocol}
\label{app:visual-shell-game-training-protocol}
All models are trained from scratch with diffusion forcing, a diffusion variant of teacher forcing, using the Wan VAE as a tokenizer.
Each five-latent-frame chunk is assigned an independently sampled noise level under a flow-matching schedule with shift $s=3$.
Causal masking is implemented with PyTorch's FlexAttention backend, and the effective batch size is $16$.
For the attention, SSM, linear-attention, and RNN families, we apply context mixing~\citep{Po2025LongContextSV} with probability $0.5$ by replacing a randomly sampled prefix with clean latents and training on the remaining noisy chunks.
LaCT and the scratchpad instead receive interleaved noisy and clean copies of each chunk.
The clean copy of chunk $t$ updates the carried state, which is then read to denoise chunk $t+1$, following prior autoregressive video applications~\citep{Zhang2025TestTimeTD}.
At inference, all models receive the two clean chunks containing the initial reveal and lowering and generate the rest autoregressively.

\begin{figure}[t]
    \centering
    \includegraphics[width=\textwidth]{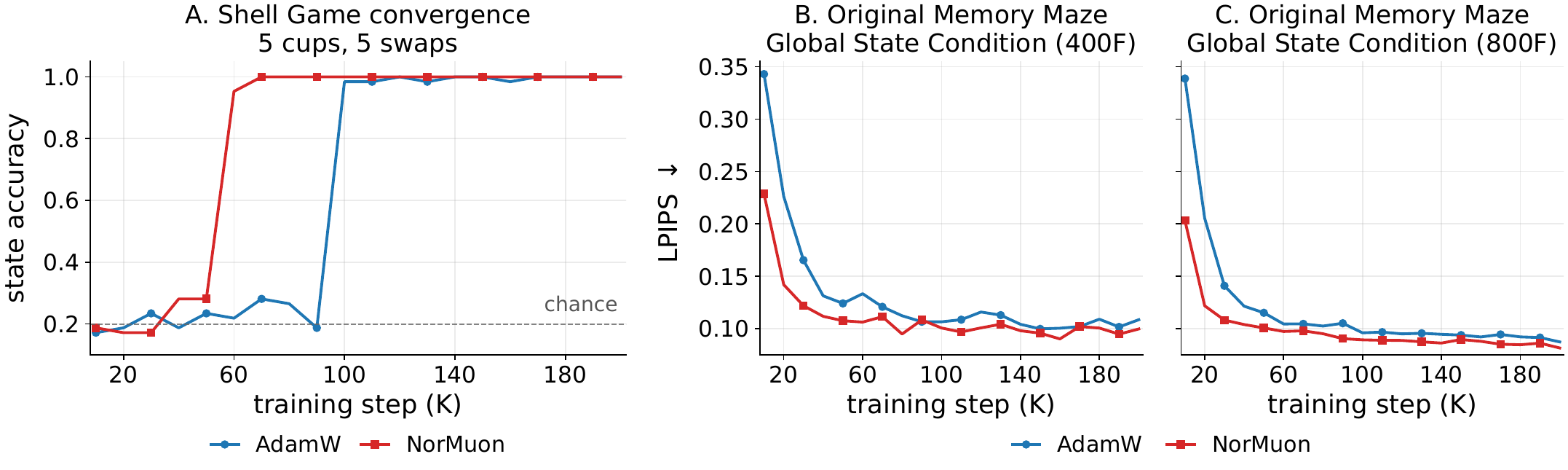}
    \caption{\textbf{NorMuon accelerates convergence.}
    All panels use Mamba2 + SWA1.
    \textbf{(A)} Shell Game state accuracy with NorMuon and AdamW.
    \textbf{(B, C)} LPIPS on the original Memory Maze with global-state conditioning at 400 and 800 frames.}
    \label{fig:optimizer-convergence}
\end{figure}

We optimize all models for $200$K steps using the NorMuon~\citep{Li2025NorMuonMM} implementation of Dion~\citep{ahn2025dion} with a base learning rate of $\eta=2\times10^{-4}$.
NorMuon updates the weight matrices of linear layers with a step size of $0.2\sqrt{\max(d_\mathrm{in},d_\mathrm{out})}\,\eta$ so that the update RMS matches AdamW, and all other parameters use AdamW with $\eta$ directly.
Training uses mixed precision through \texttt{torch.autocast}, while the RNN, LSTM, and GRU layers remain in full precision for numerical stability.
All models are trained on a single node (8 GPUs) of A100 or H100 GPUs, and a Shell Game run takes roughly one to two days on A100 GPUs, with the unoptimized LaCT and scratchpad implementations at the slower end.
NorMuon consistently converges faster than AdamW in our experiments, as shown for Mamba2 + SWA1 and the original Memory Maze in Fig.~\ref{fig:optimizer-convergence}.

\subsection{Evaluation Protocol}
\label{app:visual-shell-game-evaluation-protocol}
State accuracy is computed using a color-based ball detector on the final reveal frame.
We split the reveal frame into five equal-width regions, one per cup.
We then look for red pixels satisfying $r-\max(g,b)>0.18$ and select the region with the strongest response.
If no ball is detected, the episode is counted as incorrect, so an accuracy of $0.0$ may indicate a failure to render the ball rather than selection of the wrong cup.
We average accuracy over $64$ episodes and report $\mathrm{acc}@N$ for sequence lengths from the training horizon $N=5$ to $N=30$, against chance accuracy $1/5=0.2$, using the $200$K-step EMA weights.
SSIM, PSNR, and LPIPS are computed over the generated continuation against the ground-truth video using the same EMA checkpoints.

\section{Memory Maze Experiment Details}
\label{app:maze}

\paragraph{Setup.}
We use the Shell Game backbone families for the original Memory Maze and its two deterministic variants, each with $29$K training and $1$K validation trajectories of $1001$ frames.
Unlike the Shell Game, the RNN, Mamba2, and linear-attention backbones use spatial block sizes $[1,1,1,1,2,2,4,8]$ from bottom to top (Appendix~\ref{app:visual-shell-game-model-architecture}).
Following \citet{Po2025LongContextSV}, we train on $400$-frame clips at $128\times128$ resolution, precomputing latents with the frozen Wan2.1 VAE at $4\times$ temporal and $8\times$ spatial compression.
All backbones start from scratch with an effective batch size of $16$ and a learning rate of $2\times10^{-4}$, using AdamW on the original benchmark and NorMuon~\citep{Li2025NorMuonMM} on the deterministic datasets.
We report the EMA weights at step $200$K.

The conditioning input is either the global pose or a local action.
We encode the global pose with a two-layer MLP that projects the normalized 2D agent position and 2D heading direction to $D$, while local actions use learned embeddings for \texttt{no-op}, \texttt{forward}, \texttt{turn-left}, \texttt{turn-right}, \texttt{forward-left}, and \texttt{forward-right}.
In the local-action setting, the initial pose is provided only at the first latent step.
For either input, we pack four frame-wise conditions into each latent-frame embedding, add it to the timestep embedding, and inject it into every block through AdaLN.

\paragraph{Deterministic dataset.}
In the original MuJoCo benchmark, an agent's physical state affects how far it moves under the same discrete action.
We keep the original $9\times9$ layouts but replace this motion with grid kinematics, where a forward move advances exactly one cell and a turn rotates exactly $90$ degrees.
We used a waypoint navigation policy to move agents along shortest paths between target points, and tried our best to match the coverage statistics of the public dataset.
Like the original, each rebuilt dataset contains $29$K training and $1$K validation episodes of $1001$ frames.
In the no-collision variant, we restrict the policy to moves that succeed, so the pose can be recovered from the initial pose and action stream alone.
We also allow attempts to move into walls in the collision variant, and these leave the pose unchanged, so the model must use visual features to judge whether a forward move succeeded before updating its pose.

\paragraph{Evaluation.}
We evaluate a 220-frame continuation from 177 clean exploration frames, giving a 397-frame sequence that matches the training horizon.
For comparison with \citet{Po2025LongContextSV}, we also use 577 clean frames, closely following their evaluation with 576 context frames and a 224-frame rollout.
We call these settings 400F and 800F for simplicity.
At both lengths, the causal DiT uses its training-context SWA19 window, spanning 20 chunks or $4(5\times20)-3=397$ decoded frames.
We report SSIM, PSNR, and LPIPS using the 200K-step EMA weights, with qualitative rollouts for both conditioning schemes in Appendix~\ref{app:qualitative}.

Most recurrent variants, including LaCT, score better with the longer prefix despite training only at 400F, suggesting their learned state updates remain useful over longer observed trajectories.
On the original Memory Maze at 800F, our causal DiT reaches a PSNR of $29.25$, versus $28.8$ for the full-context causal DiT of \citet{Po2025LongContextSV}, while our Mamba2 + SWA1 reaches $32.18$, versus $28.2$ for their SSM model.
Their models receive additional 800F training, while ours train only at 400F, and differences in the DiT scaffold, latent space, and optimization may contribute to the higher PSNR.

\begin{table}[t]
\centering
\caption{\textbf{Original Memory Maze with AdamW after $200$K training steps.} Global State uses the ground-truth agent pose, while Local Action uses the discrete action sequence. Each metric scores the generated $220$-frame continuation from 177 and 577 clean context frames. Due to RoPE extrapolation issues, Causal DiT uses its training-context SWA19 window, spanning 20 chunks or 397 frames.}
\label{tab:maze-original}
\scriptsize
\setlength{\tabcolsep}{1.25pt}
\begin{tabular*}{\textwidth}{@{\extracolsep{\fill}}lcccccccccccc}
\toprule
& \multicolumn{6}{c}{Global State} & \multicolumn{6}{c}{Local Action} \\
\cmidrule(lr){2-7}\cmidrule(lr){8-13}
& \multicolumn{3}{c}{$400$ frames} & \multicolumn{3}{c}{$800$ frames}
& \multicolumn{3}{c}{$400$ frames} & \multicolumn{3}{c}{$800$ frames} \\
\cmidrule(lr){2-4}\cmidrule(lr){5-7}\cmidrule(lr){8-10}\cmidrule(lr){11-13}
Backbone
& SSIM$\uparrow$ & PSNR$\uparrow$ & LPIPS$\downarrow$
& SSIM$\uparrow$ & PSNR$\uparrow$ & LPIPS$\downarrow$
& SSIM$\uparrow$ & PSNR$\uparrow$ & LPIPS$\downarrow$
& SSIM$\uparrow$ & PSNR$\uparrow$ & LPIPS$\downarrow$ \\
\midrule
Bidirectional DiT
& 0.928 & 27.79 & 0.118 & 0.912 & 26.32 & 0.142
& \textbf{0.842} & 21.96 & \textbf{0.210} & 0.733 & 17.13 & 0.365 \\
Causal DiT (training context, SWA19)
& 0.933 & 30.39 & 0.102 & 0.938 & 29.25 & 0.094
& 0.826 & 21.85 & 0.227 & 0.846 & 22.05 & 0.194 \\
GRU + SWA1
& 0.914 & 27.23 & 0.118 & 0.918 & 27.76 & 0.113
& 0.780 & 19.02 & 0.299 & 0.824 & 21.20 & 0.232 \\
Mamba2 + SWA1
& 0.932 & 29.35 & 0.109 & \textbf{0.943} & \textbf{32.18} & 0.087
& 0.818 & 21.20 & 0.239 & 0.849 & 23.43 & 0.192 \\
GDN + SWA1
& 0.931 & 29.24 & 0.102 & 0.935 & 29.42 & 0.097
& 0.828 & 20.90 & 0.223 & 0.849 & 22.13 & 0.193 \\
GDN-neg + SWA1
& 0.926 & 27.87 & 0.111 & 0.933 & 29.40 & 0.097
& 0.827 & 20.86 & 0.225 & 0.854 & 22.81 & 0.185 \\
LaCT8 + SWA1
& \textbf{0.934} & \textbf{30.47} & \textbf{0.095} & 0.937 & 31.46 & \textbf{0.086}
& 0.827 & \textbf{22.21} & 0.230 & \textbf{0.864} & \textbf{24.29} & \textbf{0.174} \\
\bottomrule
\end{tabular*}
\end{table}
    
\begin{table}[t]
\centering
\caption{\textbf{Deterministic Memory Maze with NorMuon after $200$K training steps.} The conditioning and metric conventions follow Tab.~\ref{tab:maze-original}.}
\label{tab:maze-collision}
\scriptsize
\setlength{\tabcolsep}{1.25pt}
\begin{tabular*}{\textwidth}{@{\extracolsep{\fill}}lcccccccccccc}
\toprule
& \multicolumn{6}{c}{Global State} & \multicolumn{6}{c}{Local Action} \\
\cmidrule(lr){2-7}\cmidrule(lr){8-13}
& \multicolumn{3}{c}{$400$ frames} & \multicolumn{3}{c}{$800$ frames}
& \multicolumn{3}{c}{$400$ frames} & \multicolumn{3}{c}{$800$ frames} \\
\cmidrule(lr){2-4}\cmidrule(lr){5-7}\cmidrule(lr){8-10}\cmidrule(lr){11-13}
Backbone
& SSIM$\uparrow$ & PSNR$\uparrow$ & LPIPS$\downarrow$
& SSIM$\uparrow$ & PSNR$\uparrow$ & LPIPS$\downarrow$
& SSIM$\uparrow$ & PSNR$\uparrow$ & LPIPS$\downarrow$
& SSIM$\uparrow$ & PSNR$\uparrow$ & LPIPS$\downarrow$ \\
\midrule
\multicolumn{13}{l}{\textit{No collision}} \\
\midrule
Causal DiT (training context, SWA19)
& 0.945 & 30.51 & \textbf{0.075} & 0.948 & 32.09 & 0.072
& 0.934 & 29.02 & 0.087 & 0.935 & 30.42 & 0.083 \\
GRU + SWA1
& 0.940 & 31.48 & 0.082 & 0.944 & 32.56 & 0.077
& 0.883 & 22.73 & 0.177 & 0.899 & 24.15 & 0.145 \\
Mamba2 + SWA1
& 0.942 & 31.43 & 0.083 & 0.949 & 33.53 & 0.069
& 0.928 & 29.75 & 0.097 & 0.936 & 31.56 & 0.082 \\
GDN + SWA1
& 0.942 & 30.20 & 0.080 & 0.948 & 33.22 & 0.071
& 0.929 & 30.37 & 0.094 & 0.936 & 31.96 & 0.082 \\
GDN-neg + SWA1
& 0.942 & 30.13 & 0.080 & 0.948 & 33.25 & 0.071
& 0.932 & 30.83 & 0.089 & 0.938 & 32.16 & 0.080 \\
LaCT8 + SWA1
& \textbf{0.946} & \textbf{32.25} & 0.077 & \textbf{0.952} & \textbf{33.94} & \textbf{0.065}
& \textbf{0.943} & \textbf{31.47} & \textbf{0.081} & \textbf{0.952} & \textbf{33.57} & \textbf{0.066} \\
\midrule
\multicolumn{13}{l}{\textit{With collision}} \\
\midrule
Causal DiT (training context, SWA19)
& 0.911 & 27.84 & 0.112 & 0.923 & 28.69 & 0.091
& 0.859 & 24.62 & 0.173 & 0.884 & 24.84 & 0.137 \\
GRU + SWA1
& 0.892 & 25.65 & 0.140 & 0.911 & 28.73 & 0.108
& 0.765 & 17.57 & 0.331 & 0.792 & 18.40 & 0.285 \\
Mamba2 + SWA1
& 0.910 & 27.63 & 0.113 & 0.926 & 30.84 & 0.085
& 0.827 & 21.14 & 0.230 & 0.863 & 24.54 & 0.171 \\
GDN + SWA1
& 0.907 & 24.82 & 0.121 & 0.929 & 30.90 & 0.083
& 0.853 & 23.76 & 0.187 & 0.878 & 26.39 & 0.149 \\
GDN-neg + SWA1
& 0.908 & 25.11 & 0.119 & 0.929 & 30.58 & 0.085
& 0.863 & 24.98 & 0.171 & 0.895 & \textbf{28.24} & 0.122 \\
LaCT8 + SWA1
& \textbf{0.917} & \textbf{28.42} & \textbf{0.103} & \textbf{0.937} & \textbf{32.05} & \textbf{0.075}
& \textbf{0.878} & \textbf{25.16} & \textbf{0.158} & \textbf{0.906} & \textbf{28.24} & \textbf{0.116} \\
\bottomrule
\end{tabular*}
\end{table}
    
\paragraph{Results.}
Tabs.~\ref{tab:maze-original} and~\ref{tab:maze-collision} report results under global-state and local-action conditioning.
Prior long-context Memory Maze studies primarily use global-state conditioning, which provides the exact agent pose at every frame~\citep{Po2025LongContextSV,Chen2025RecurrentAD}.
Each observation is therefore effectively tagged by its pose, allowing the recurrent state or KV cache to act as a pose-indexed visual memory.
Mamba2 and LaCT perform well in this setting, supporting their ability to store and retrieve visual information over long trajectories.
Local-action conditioning provides the initial pose followed by discrete controls, so the model must also estimate its current pose to retrieve the corresponding visual information.
The task therefore combines state tracking with visual retrieval rather than retrieval alone.
On the original Memory Maze, we found that all models perform substantially worse under local-action conditioning.
Further inspection suggested that the models accumulate localization error over long trajectories because the MuJoCo dynamics can produce different displacements for the same discrete action.
Although wall collisions are rare in the released trajectories, these dynamics make accurate dead reckoning from the action sequence difficult.

We designed the deterministic variants to isolate whether this drop comes from local-action conditioning itself or from drift in the MuJoCo dynamics.
Without collisions, each action applies a fixed translation or rotation, making the pose a deterministic function of the initial pose and action sequence.
Pose tracking can then be expressed as a prefix-sum-style composition of fixed action-induced transformations.
The global-state and local-action results become much closer in this setting, indicating that local-action conditioning itself is not the main source of the original gap.
At 800F, LaCT reaches the same SSIM of 0.952 under both global-state and local-action conditioning, essentially closing the gap between the two.

Allowing collisions makes the transition also depend on what the model sees. 
A blocked forward move leaves the pose unchanged, so the action sequence alone no longer determines the trajectory, and the model has to judge from the frames whether each move succeeded before updating its state. 
This is no longer the action-only prefix accumulation of the collision-free variant and all backbones' performance drops relative to their no-collision variant, with the drop being larger under local-action conditioning.
Global-state conditioning suffers too because wall contacts make the visual transitions harder, but local-action conditioning on top of that has to resolve their effect on pose.

LaCT is the strongest backbone across most 800F settings, particularly under local-action conditioning.
We hypothesize that LaCT benefits from its expressive nonlinear fast-weight update, which may better represent transitions that combine actions with visual evidence.
In contrast, widening the linear transition spectrum to include negative eigenvalues yields no consistent gain, with GDN-neg performing like GDN.
This suggests Memory Maze places different structural demands on the state transition and does not show the clear benefit from negative eigenvalues seen in the Shell Game.

\section{3D Block World Experiment Details}
\label{app:blockworld}
\paragraph{Setup.}
We use the textured 3D Block World of \citet{Lillemark2026FlowEW}, a MiniWorld-based room in which blocks move on their own even while the agent's camera is turned away.
Episodes contain 500 frames at $128\times128$ resolution, with the agent controlled by \texttt{no-op}, \texttt{forward}, \texttt{turn-left}, and \texttt{turn-right}.
Each turn changes the view by exactly 90 degrees, and we follow the Memory Maze local-action protocol by providing the normalized initial pose only at the first latent step.

We found that the Wan VAE's $4\times$ temporal compression handles these abrupt turns poorly and use FloWM's framewise MAE-ViT-based image VAE instead.
Its 8-channel latents have $16\times$ spatial compression and no temporal compression, so each action corresponds to one latent frame.
We group five latent frames into an autoregressive chunk and train all backbones from scratch on 160-frame clips for 200K steps, using NorMuon with a learning rate of $2\times10^{-4}$ and an effective batch size of 32.
As on the Shell Game, the models train with a clean prefix covering a random half to all of the leading chunks, while LaCT interleaves the clean and noisy copies of each chunk and uses no such prefix (Appendix~\ref{app:visual-shell-game-training-protocol}).

\paragraph{Evaluation.}
We provide each model with a clean 80-frame prefix and generate continuations of either 80 or 240 frames, yielding sequences of 160 frames at the training length and 320 frames beyond it.
Using the 200K-step EMA weights, we score the generated continuation with SSIM, PSNR, and LPIPS.

\paragraph{Results.}
At roughly 200M parameters, all backbones model the training-length rollouts reasonably well, and most remain visually coherent at 320 frames.
Among the models restricted to SWA1, LaCT performs substantially better than the other recurrent backbones.
At 320 frames, LaCT reaches a PSNR of $29.36$ and an LPIPS of $0.0288$, compared with $21.32$ and $0.1060$ for Mamba2.
LaCT is trained without an explicit state target, and its architecture does not impose a rule for tracking out-of-view objects.
We suspect that its nonlinear fast-weight update can represent information about these objects more flexibly.
By comparison, GDN performs similarly to Mamba2, and widening its transition spectrum with GDN-neg yields little additional benefit.

In our experiments, increasing the local attention window has a larger effect than changing the recurrent update.
For Mamba2, expanding attention from SWA1 to SWA10 improves the 320-frame PSNR from $21.32$ to $32.29$ and reduces LPIPS from $0.1060$ to $0.0167$ (Tab.~\ref{tab:blockworld-tex-mamba-window}).
We find that longer recent context is particularly helpful here because framewise encoding and abrupt 90-degree turns make adjacent observations less redundant.
The improvement peaks at SWA10, with performance declining again at SWA15 and SWA20.

We also note that the windowed baselines are strong relative to the results reported by \citet{Lillemark2026FlowEW}.
Their textured Block World evaluation uses 70 context frames followed by 210 generated frames, where FloWM, DFoT, and DFoT-SSM reach PSNR scores of $30.33$, $20.43$, and $19.15$.
In our slightly longer setting with 80 context frames and a 240-frame continuation, the roughly 200M-parameter Causal SWA5, Mamba2 with SWA10, and LaCT with SWA1 reach $37.60$, $32.29$, and $29.36$.
These results suggest that much of the task can be addressed through backbone scaling and sufficient recent context.
The open-room layout may contribute to this behavior because moving objects often remain visible or return within the attention window.

\begin{table}[h]
\centering
\caption{\textbf{Textured 3D Block World backbone comparison under local-action conditioning after $200$K training steps.} All metrics use the EMA weights. SWA$w$ attends to the current and $w$ previous chunks, with five framewise latents per chunk.}
\label{tab:blockworld-tex-backbones}
\small
\setlength{\tabcolsep}{4.5pt}
\begin{tabular*}{\textwidth}{@{\extracolsep{\fill}}lccccccc}
\toprule
& & \multicolumn{3}{c}{$160$ frames} & \multicolumn{3}{c}{$320$ frames} \\
\cmidrule(lr){3-5}\cmidrule(lr){6-8}
Backbone & $w$
& SSIM$\uparrow$ & PSNR$\uparrow$ & LPIPS$\downarrow$
& SSIM$\uparrow$ & PSNR$\uparrow$ & LPIPS$\downarrow$ \\
\midrule
Causal DiT (training context, SWA31) & 31 & \textbf{0.9944} & \textbf{42.03} & \textbf{0.0058} & 0.9630 & 24.98 & 0.0495 \\
Causal DiT (SWA5) & 5 & 0.9922 & 40.02 & 0.0088 & \textbf{0.9907} & \textbf{37.60} & \textbf{0.0111} \\
Causal DiT (SWA1) & 1 & 0.9376 & 22.02 & 0.0877 & 0.9104 & 20.21 & 0.1284 \\
Mamba2 + SWA1 & 1 & 0.9473 & 23.88 & 0.0745 & 0.9255 & 21.32 & 0.1060 \\
GDN + SWA1 & 1 & 0.9469 & 23.96 & 0.0747 & 0.9250 & 21.52 & 0.1067 \\
GDN-neg + SWA1 & 1 & 0.9478 & 24.42 & 0.0723 & 0.9260 & 21.36 & 0.1037 \\
LaCT8 + SWA1 & 1 & 0.9908 & 38.26 & 0.0114 & 0.9791 & 29.36 & 0.0288 \\
\bottomrule
\end{tabular*}
\end{table}

\begin{table}[h]
    \centering
    \caption{\textbf{Mamba2 SWA sweep on Textured 3D Block World under local-action conditioning after $200$K training steps.} The metric and SWA conventions follow Tab.~\ref{tab:blockworld-tex-backbones}.}
    \label{tab:blockworld-tex-mamba-window}
    \small
    \setlength{\tabcolsep}{7pt}
    \begin{tabular*}{\textwidth}{@{\extracolsep{\fill}}ccccccc}
    \toprule
    & \multicolumn{3}{c}{$160$ frames} & \multicolumn{3}{c}{$320$ frames} \\
    \cmidrule(lr){2-4}\cmidrule(lr){5-7}
    $w$
    & SSIM$\uparrow$ & PSNR$\uparrow$ & LPIPS$\downarrow$
    & SSIM$\uparrow$ & PSNR$\uparrow$ & LPIPS$\downarrow$ \\
    \midrule
    1 & 0.9473 & 23.88 & 0.0745 & 0.9255 & 21.32 & 0.1060 \\
    2 & 0.9738 & 29.58 & 0.0358 & 0.9476 & 23.62 & 0.0744 \\
    3 & 0.9885 & 35.98 & 0.0150 & 0.9762 & 28.87 & 0.0338 \\
    5 & 0.9921 & 39.40 & 0.0093 & 0.9837 & 30.29 & 0.0234 \\
    10 & 0.9926 & 40.21 & 0.0083 & \textbf{0.9870} & \textbf{32.29} & \textbf{0.0167} \\
    15 & \textbf{0.9940} & \textbf{40.67} & \textbf{0.0066} & 0.9746 & 27.07 & 0.0332 \\
    20 & 0.9939 & 40.57 & 0.0067 & 0.9503 & 23.27 & 0.0645 \\
    \bottomrule
    \end{tabular*}
\end{table}

\section{Additional Discussion and Limitations}
\label{app:discussion}

\noindent\textbf{Global State Conditioning vs Local Action Conditioning.}
Video world models often take global camera poses or prescribed trajectories as input~\citep{Ren2025Gen3C3W,Huang2025VoyagerLA,Wu2025GeometryFM,Nam2026WorldCamIA,Yi2026WorldKVEW}, which allows precise viewpoint control and retrieval of earlier views by pose.
For interactive use, a calibrated controller can translate keyboard and mouse inputs into these camera motions~\citep{Nam2026WorldCamIA}, so the model is given its camera state without having to localize itself.
Raw discrete actions are cheaper to obtain, but then the model has to infer how each control changes its global state.
Although there are some local action-conditioned models in interactive gaming domain~\citep{He2025MatrixGame2A,Wang2026MatrixGame3R,Savva2026SolarisBA}, we still think pose recovery from an initial pose and subsequent local actions has received little direct study.
In our MuJoCo runs, no local-action-conditioned backbone matched global-state fidelity.
The same discrete action can produce different displacements, and errors in the accumulated pose estimates lead to drift over a long rollout.
We believe recovering the physics engine's state from observations and discrete local actions alone, and correcting its drift over long horizons, remains an open problem.

\noindent\textbf{State Carriers and Learnability.}
An LLM can append a token that snapshots the current state and read it back at the next step~\citep{Merrill2024TheEP}, and supervised traces or sequence-level rewards can teach that protocol.
Pure video training provides no such target.
In our setup, flow matching trains each chunk toward a reconstruction latent of the frozen Wan VAE, which carries no label for the hidden state, so generated frames cannot serve as a scratchpad (Appendix~\ref{app:no-scratchpad}).
Large systems are beginning to split the problem accordingly, with Cosmos 3 routing reasoning through a dedicated autoregressive tower while a separate diffusion generator renders from its representations~\citep{Aditi2026Cosmos3O}.
Keeping the state inside the video backbone is what our recurrent variants test, but we often found expressivity alone is not enough.
The clearest case is the nonlinear RNNs, which solve the toy $S_5$ task in a single layer, yet inside the video stack the vanilla RNN never fits even the training length, while the GRU and LSTM fit it but fall toward chance beyond it.
Their Shell Game accuracies at $N=5/10/20/30$ are $0.22/0.14/0.20/0.22$ for the RNN, $1.00/0.30/0.23/0.25$ for the GRU, and $1.00/0.17/0.19/0.22$ for the LSTM.
LaCT (SwiGLU) with a single fast-weight head fails to extrapolate even though it carries the largest state, while more heads track better with less state, pointing to factorized updates or easier optimization rather than raw capacity.
The benefit has limits of its own.
For example, raising the LaCT head count from 8 to 48 lowered visual fidelity on the deterministic Memory Maze, suggesting that maintaining video quality needs a sufficient head dimension, so we keep LaCT8 as the reference throughout.
The inner architecture, update loss, normalization, and update schedule all shape whether a formally expressive fast-weight model learns a useful transition, so a failing run may reflect a learning difficulty rather than a representational limit.

\noindent\textbf{Effect of Chunk Size in LaCT.}
LaCT performs one serial fast-weight update per chunk, so a sequence of length $L$ processed in chunks of size $c$ receives roughly $L/c$ serial updates, and the swaps inside a chunk must be composed by the fixed-depth chunk processor before the state is written (Sec.~\ref{app:ttt-linear-attention}).
As the toy depth sweep in Fig.~\ref{fig:toy-s3-s5-exp} shows, a Transformer needs more layers to compose a longer swap chain, so larger chunks push work from the recurrence back onto backbone depth.
We tried a chunk-size ablation for TTT layers in the toy setting, but small chunks unroll the fast-weight update into a very long gradient chain, and training became too slow and unstable to finish the sweep.
The runs we could finish still pointed the expected way, with smaller chunks solving the task at lower depth.
On video the same cost stretches across hundreds of latent frames, so we did not attempt a controlled sweep there.
Matrix-state linear attention is easier to move along this axis, since its outer-product writes can be defined at token, frame, or group granularity while keeping optimized parallel kernels, though we did not test these granularities here. 
Nonlinear fast weights have no such kernels, and in our runs LaCT and the scratchpad train markedly slower than the attention and linear-attention baselines, both because interleaving clean and noisy copies of each chunk doubles the effective sequence length and because the carried state serializes training across chunks. 
Nonetheless, they are still the easier tool for research, since the apply and update operations are written directly in PyTorch and can be rearranged or replaced.

\noindent\textbf{Limitations.}
The Shell Game maps directly onto $S_5$ state tracking, but the more general settings in Sec.~\ref{sec:beyond} have no such clear correspondence.
Part of their state must be corrected from observations, often without the relevant changes being annotated by action inputs, and we do not yet know which state-tracking abstraction best describes these settings.
Our experiments test length extrapolation on synthetic tasks at one model scale, around 200M parameters, with one flow-matching recipe.
Whether the same mechanisms work on real-world video or at the scale of modern video world models remains untested.
Nonetheless, video models remain a scalable way to learn world dynamics, and we believe understanding and learning the structure of the diverse state updates and implicit states they require is a promising research direction.

\section{Qualitative Samples}
\label{app:qualitative}

\begin{figure}[h]
\centering
\includegraphics[width=\textwidth]{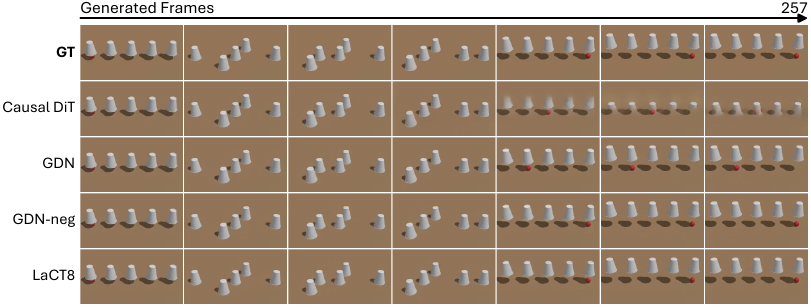}
\caption{\textbf{Shell Game rollout at twice the training length.}
Models trained at $N{=}5$ are rolled out over $N{=}10$ swaps.
Rows are the ground truth and four backbones, where the causal DiT keeps its training-context SWA7 window and the recurrent backbones use SWA1.
At the final reveal, GDN-neg and LaCT8 land on the same cup as the ground truth, while the causal DiT and GDN do not.
Every causal window misses the cup, and frames get rougher as the window widens and drift once it exceeds the training context.
A training episode has eight chunks, so wider windows see fewer truncated contexts during training, and SWA7 never sees one.
Beyond $N{=}5$ the window must drop the earliest chunks, including the initial reveal, which SWA7 never learned to handle, so it has the lowest PSNR among the causal DiTs in Fig.~\ref{fig:shellgame_main} and blurs the later frames here.}
\label{fig:shellgame-qualitative}
\end{figure}

\begin{figure}[h]
\centering
\includegraphics[width=0.77\textwidth]{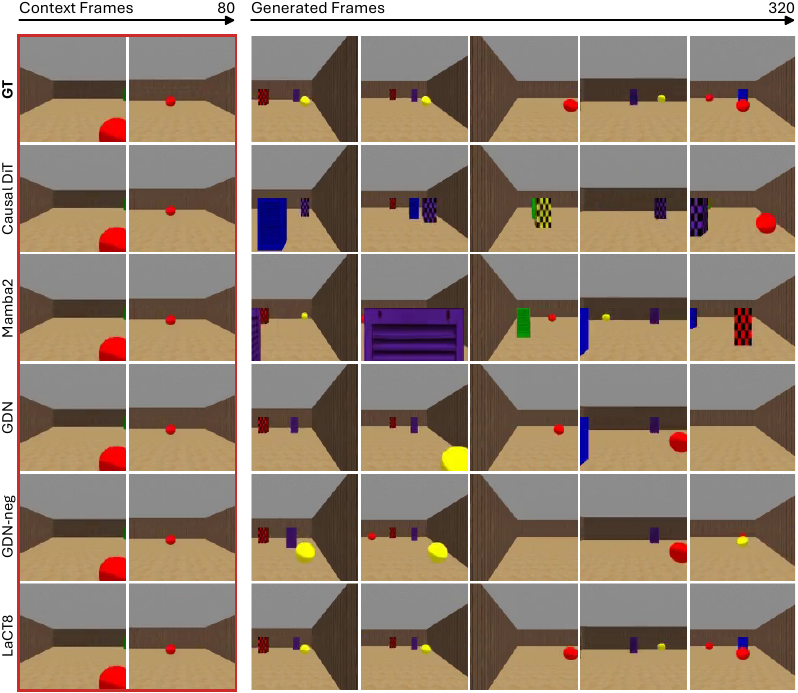}
\caption{\textbf{Textured 3D Block World under local-action conditioning.}
The rollout uses an 80-frame clean prefix and a 240-frame continuation (Tab.~\ref{tab:blockworld-tex-backbones}).
All rows use SWA1, so they differ only in the recurrent update.
LaCT8 keeps the blocks and the ball near their positions in the ground truth, while the other backbones render a coherent room whose contents have drifted.}
\label{fig:blockworld-qualitative}
\end{figure}

\begin{figure}[p]
\centering
\includegraphics[width=0.9\textwidth]{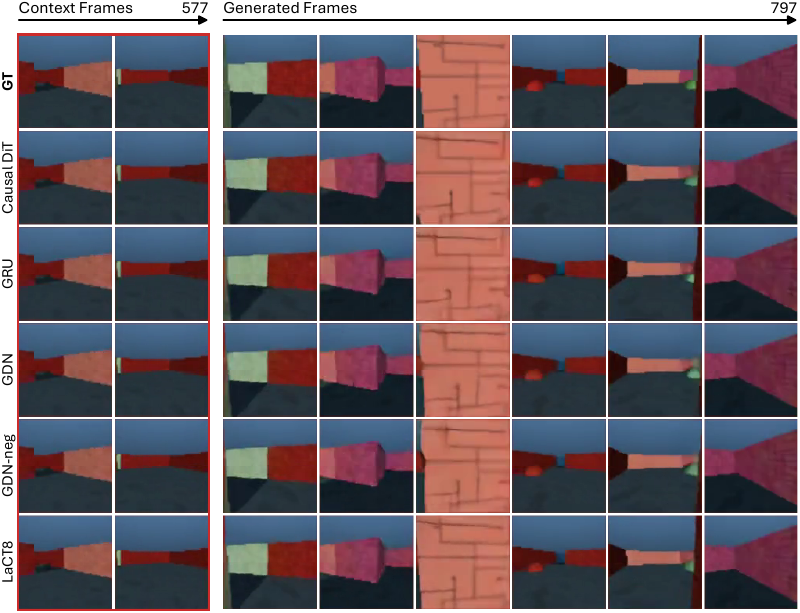}
\caption{\textbf{Original Memory Maze under global-state conditioning.}
Rows are the ground truth and the backbones of Tab.~\ref{tab:maze-original} (original Memory Maze), where the causal DiT uses its training-context SWA19 window and the recurrent backbones use SWA1.}
\label{fig:maze-orig-global}
\end{figure}

\begin{figure}[p]
\centering
\includegraphics[width=0.9\textwidth]{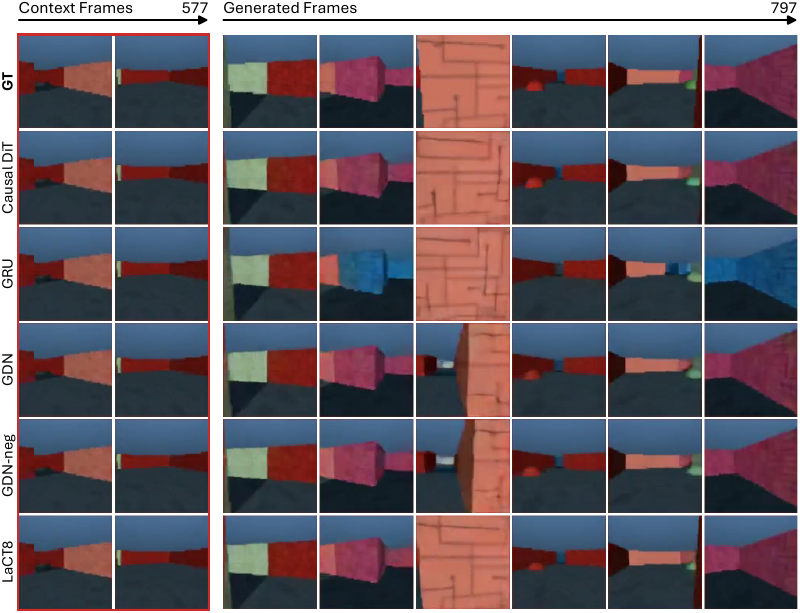}
\caption{\textbf{Original Memory Maze under local-action conditioning.}
The episode and frames are those of Fig.~\ref{fig:maze-orig-global}.}
\label{fig:maze-orig-local}
\end{figure}

\begin{figure}[p]
\centering
\includegraphics[width=0.9\textwidth]{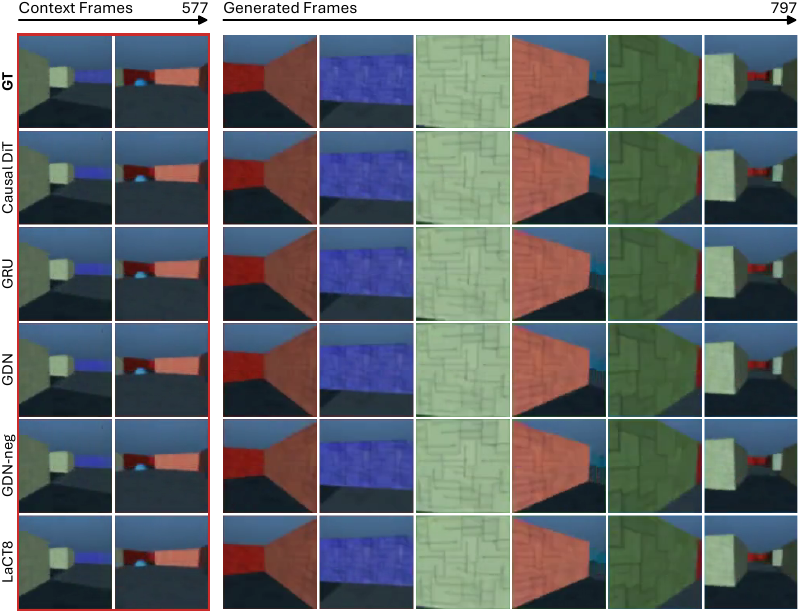}
\caption{\textbf{Deterministic Memory Maze without collisions, global state.}
Conventions follow Fig.~\ref{fig:maze-orig-global}, with the backbones of Tab.~\ref{tab:maze-collision} (deterministic Memory Maze without collisions).}
\label{fig:maze-nocoll-global}
\end{figure}

\begin{figure}[p]
\centering
\includegraphics[width=0.9\textwidth]{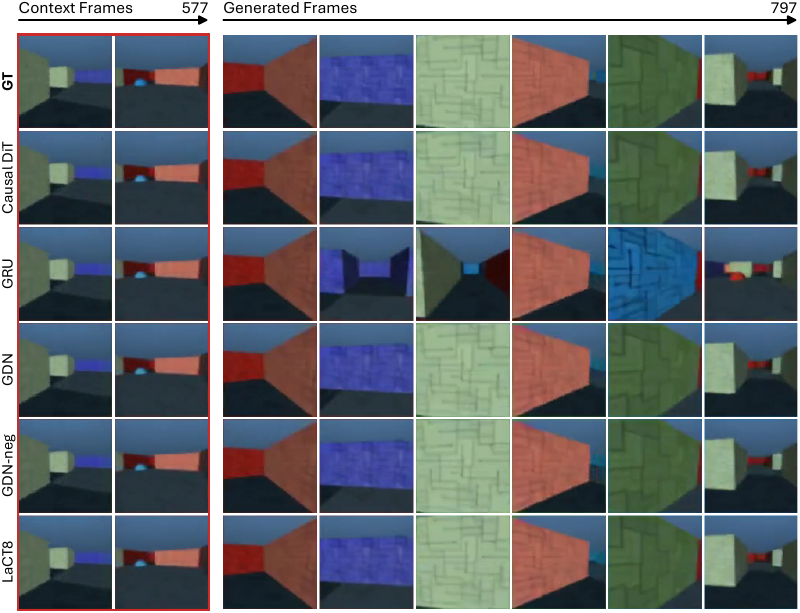}
\caption{\textbf{Deterministic Memory Maze without collisions, local action.}
Pose is a pure function of the action stream in this variant, and every backbone except the GRU stays close to the ground truth through the rollout.}
\label{fig:maze-nocoll-local}
\end{figure}

\begin{figure}[p]
\centering
\includegraphics[width=0.9\textwidth]{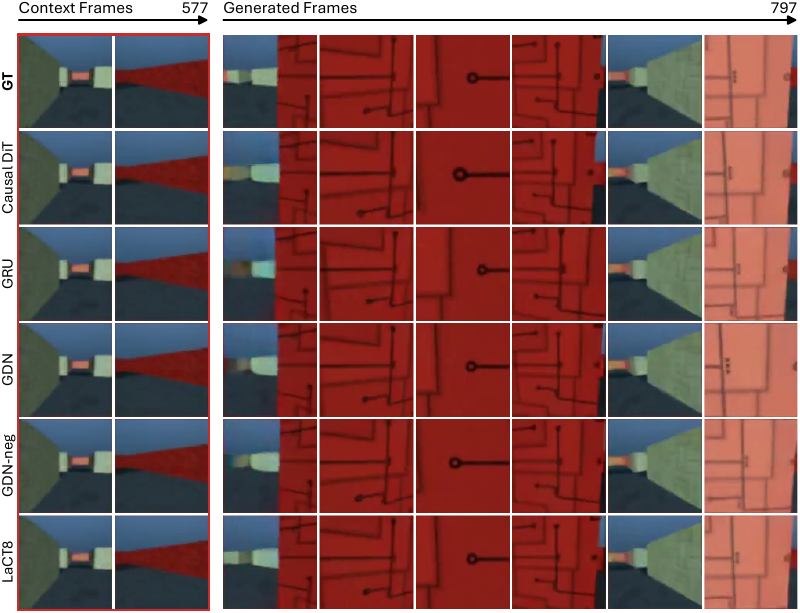}
\caption{\textbf{Deterministic Memory Maze with collisions, global state.}
Conventions follow Fig.~\ref{fig:maze-orig-global}, with the backbones of Tab.~\ref{tab:maze-collision} (deterministic Memory Maze with collisions).}
\label{fig:maze-coll-global}
\end{figure}

\begin{figure}[p]
\centering
\includegraphics[width=0.9\textwidth]{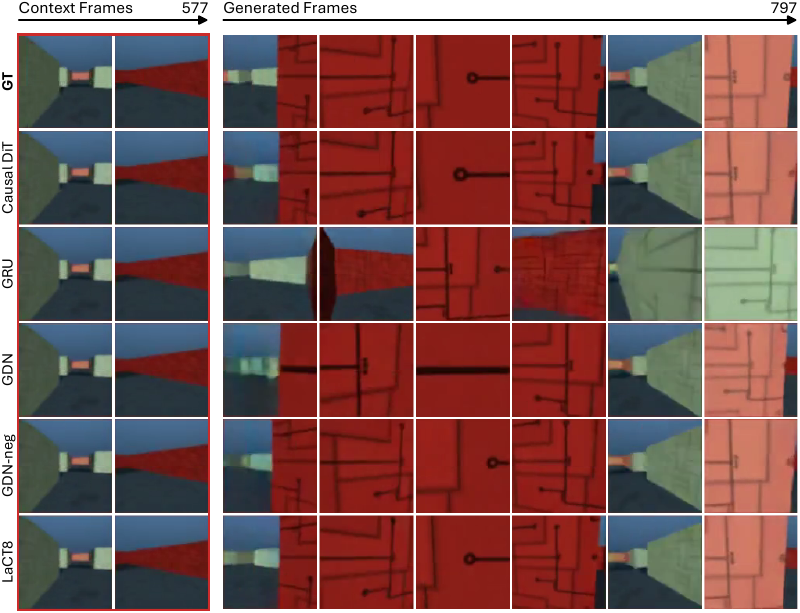}
\caption{\textbf{Deterministic Memory Maze with collisions, local action.}
A blocked move leaves the pose unchanged, so the update depends on the layout the model cannot see.}
\label{fig:maze-coll-local}
\end{figure}

We show generated rollouts for the three experiments in the paper. 
Each figure follows one episode, and all rows share the same context frames before generating the rest independently. 
Fig.~\ref{fig:shellgame-qualitative} shows the Shell Game and Fig.~\ref{fig:blockworld-qualitative} shows the textured 3D Block World. 
For the Memory Maze, Figs.~\ref{fig:maze-orig-global} to~\ref{fig:maze-coll-local} show the original and deterministic datasets (without and with collision) under the 800F evaluation of Appendix~\ref{app:maze}, with both conditioning schemes on the same episode of each dataset.

\end{document}